\pdfoutput=1 
\documentclass{article}
\usepackage{graphicx}
\usepackage{algorithm}
\usepackage{algorithmicx}
\usepackage{amsmath,amssymb,amsfonts}
\usepackage{multicol}
\usepackage{multirow}
\usepackage{bm}
\usepackage{amsthm}
\usepackage{algpseudocode}
\usepackage{mathrsfs}
\usepackage{geometry}
\usepackage{indentfirst}
\usepackage{pifont}
\usepackage{ulem}
\usepackage{color}
\usepackage{caption}
\usepackage{subcaption}
\usepackage{booktabs}
\usepackage{hyperref}
\usepackage{appendix}  
\usepackage{authblk}

\begin{document}
	\title{Gradient Episodic Memory with a Soft Constraint for Continual Learning}
	
	\author[a]{Guannan Hu}
	\author[a]{Wu Zhang}
	\author[b]{Hu Ding}
	\author[a]{Wenhao Zhu \thanks{Corresponding author: whzhu@shu.edu.cn}}
	\affil[a]{School of Computer Engineering and Science, Shanghai University, Shanghai, CHN}
	\affil[b]{Shanghai Institute of Applied Mathematics and Mechanisc, Shanghai, CHN}

	\maketitle

\begin{abstract}
Catastrophic forgetting in continual learning is a common destructive phenomenon in gradient-based neural networks that learn sequential tasks, and it is much different
%Editor: On the line below, please ensure that the intended meaning has been maintained in this edit.
from forgetting in humans,
who can learn and accumulate knowledge throughout their whole lives.
Catastrophic forgetting is the fatal shortcoming of a large decrease in performance on previous tasks when the model is learning a novel task.
To alleviate this problem, the model should have the capacity to learn new knowledge and preserve learned knowledge. We propose an  average gradient episodic memory (A-GEM) with a soft constraint $\epsilon \in [0, 1]$, which is a balance factor between learning new knowledge and preserving learned knowledge; our method is called gradient episodic memory with a soft constraint $\epsilon$ ($\epsilon$-SOFT-GEM).
$\epsilon$-SOFT-GEM outperforms A-GEM and several continual learning benchmarks in a single training epoch; additionally, it has state-of-the-art average accuracy and efficiency for computation and memory, like A-GEM, and provides a better trade-off between the stability of preserving learned knowledge and the plasticity of learning new knowledge.
\end{abstract}

% keywords can be removed

\section{Introduction}
\label{Introduction}
In general, humans observe data as a sequence and seldom observe the samples twice; otherwise, they can learn and accumulate knowledge of new data throughout their whole lives.
Unlike humans, artificial neural networks (ANNs), which are inspired by biological neural systems, suffer from catastrophic forgetting \cite{McClelland1995Why,Mccloskey1989CatastrophicII,Ratcliff1990Connectionnist}, whereby learned knowledge is disrupted when a new task is being learned.

Continual learning \cite{Hassabis2017Neuroscience,Ring1994Continual,Thrun1995Lifelong} aims to alleviate catastrophic forgetting in ANNs. The key to continual learning is that the model handles the data individually and preserves the knowledge of previous tasks without storing all the data from previous tasks. With continual learning, the model has the potential to learn novel tasks quickly if it can consolidate the previously acquired knowledge. Unfortunately, in the
%Editor: On the line below, please ensure that the intended meaning has been maintained in this edit.
most common approaches,
the model cannot learn new knowledge about previous tasks when acquiring new knowledge of new tasks in order to alleviate catastrophic forgetting. For example, EWC\cite{Kirkpatrick2017OvercomingCF}, PI \cite{Zenke2017ContinualLT}, RWALK \cite{Chaudhry2018RiemannianWF} and MAS \cite{Aljundi2018MemoryAS}, which use regularization to slow down learning with weights that correlate with previously acquired knowledge, resist decreasing the performance on previous tasks and cannot acquire new knowledge fast. The assumption of gradient episodic memory (GEM) \cite{LopezPaz2017GradientEM} and average gradient episodic memory (A-GEM) \cite{Chaudhry2018EfficientLL} is that the model guarantees to avoid
%Editor: On the line below, please ensure that the intended meaning has been maintained in this edit.
increasing  the loss over episodic memory
when the model updates the gradient, which has the same shortcoming.

Catastrophic forgetting can be alleviated if the model can acquire novel knowledge about previous tasks when learning new tasks.
Building on GEM and A-GEM, we assume that the model not only maintains the loss over episodic memory, preventing it from increasing, but actually decreases the loss to acquire novel knowledge of experiences that are representative of the previous tasks.
To achieve this goal, the optimizer of the model should guarantee that the angle between the gradient of samples from episodic memory and the updated gradient is less than $90^{\circ}$.
Based on the idea above, we introduce a soft constraint $\epsilon \in [0, 1]$, which is a balance between forgetting old tasks (loss over previous tasks that are represented by episodic memory) and learning new tasks (loss over new tasks), and propose a variant of A-GEM with a soft constraint $\epsilon$, called $\epsilon$-SOFT-GEM, which is a combination of episodic memory and optimization constraints.
Additionally, we introduce an intuitive idea, average A-GEM (A-A-GEM), in which the updated gradient is the average of the gradient of samples from episodic memory and the gradient of new samples from learning task, and the angle between the gradient of the samples from episodic memory and the updated gradient must be no more than $90^{\circ}$.

We evaluate $\epsilon$-SOFT-GEM, A-A-GEM and several representative baselines on a variety of sequential learning tasks on the metrics of the stability and plasticity of the model. Our experiments demonstrate that $\epsilon$-SOFT-GEM achieves better performance than A-GEM with almost the same efficiency in terms of computation and memory; meanwhile, $\epsilon$-SOFT-GEM outperforms other common continual learning benchmarks in a single training epoch.

\section{Related Work}
\label{Related Works}
The term catastrophic forgetting was first introduced by \cite{Mccloskey1989CatastrophicII}, who claimed that catastrophic forgetting is a fundamental limitation of neural networks and a downside of their high generalization ability.
The cause of catastrophic forgetting is that ANNs are based on concurrent learning, where the whole population of the training samples is presented and trained as a single and complete entity; therefore, alterations to the parameters of ANNs using back-propagation lead to catastrophic forgetting when training on new samples.

Several works have described
%Editor: On the line below, please ensure that the intended meaning has been maintained in this edit.
destructive consequences of catastrophic forgetting in
sequential learning and provided a few primitive solutions, such as employing experience replay with all previous data or subsets of previous data \cite{Robins1993Catastrophic,Robins1995Catastrophic}.

Other works focus on training individual models or sharing structures to alleviate catastrophic forgetting.
A progressive neural network (PROGNN) \cite{Rusu2016ProgressiveNN} has been proposed to train individual models on each task, retain a pool of pretrained models and learn lateral connections from these to extract useful features for new tasks, which eliminates forgetting altogether but requires growing the network after each task and can cause the architecture complexity to increase with the number of tasks.
DEN \cite{Yoon2018Lifelong} can learn a compact overlapping knowledge-sharing structure among tasks.
PROGNN and DEN require the number of parameters to be constantly increased and thus lead to a huge and complex model.

Many works focus on optimizing network parameters on new tasks while minimizing alterations to the consolidated weights on previous tasks.
It is suggested that regularization methods, such as dropout\cite{Hinton2012Improving}, L2 regularization and activation functions \cite{Glorot2011Deep,Ian2013Maxout}, help to reduce forgetting for previous tasks \cite{Goodfellow2014AnEI}.
Furthermore, EWC \cite{Kirkpatrick2017OvercomingCF} uses a Fisher information matrix-based regularization to slow down learning on network weights that correlate with previously acquired knowledge.
PI \cite{Zenke2017ContinualLT} employs the path integrals of loss derivatives to slow down learning on weights that are important for previous tasks.
MAS \cite{Aljundi2018MemoryAS} accumulates an importance measure for each parameter of the network and penalizes the important parameters.
RWALK \cite{Chaudhry2018RiemannianWF} introduces a distance in the Riemannian manifold as a means of regularization.
The regularization methods resist decreasing the performance on previous tasks and learn new tasks slowly.

Episodic memory can store previously seen samples and replay them; iCarl \cite{Rebuffi2017iCaRLIC} replays the samples in episodic memory, while GEM \cite{LopezPaz2017GradientEM} and A-GEM \cite{Chaudhry2018EfficientLL} use episodic memory to make future gradient update.
However, choosing samples from previous tasks is challenging since it requires knowing how many samples to store and how the samples represent the tasks.
The experience replay strategies proposed in \cite{Chaudhry2019Continual} are reservoir sampling \cite{Riemer2019Learning}, ring buffer \cite{LopezPaz2017GradientEM}, k-means and mean of features \cite{Rebuffi2017iCaRLIC}.

\section{Gradient Episodic Memory with a Soft Constraint}
\label{softgem}
For $\epsilon$-SOFT-GEM, the learning protocol is described in \cite{Chaudhry2018EfficientLL}, and the sequential learning task is divided into two ordered sequential streams $D^{CV} = \{D_{1}, ..., D_{T^{CV}}\}$ and $D^{EV}=\{D_{T^{CV}+1}, ..., D_{T}\}$, where $D_{k}=\{(\mathbf{x}_{i}^{k}, t_{i}^{k}, y_{i}^{k})_{i=1}^{n_{k}}\}$ is the dataset of the $k$-th task, $T^{CV} <T$. $D^{CV}$ is the stream of datasets used in cross-validation to select the hyperparameters of the model, and $D^{EV}$ is the stream of datasets used for training and evaluation.

\vspace*{-0.05in}
\subsection{GEM}
\vspace*{-0.05in}

In this section, we review GEM, which is a model for continual learning with an episodic memory $\mathcal{M}_{k}$ for each task $k$, which stores a subset of the observed examples or embedding features from task $k$. GEM defines the loss over $\mathcal{M}_{k}$ as:
\begin{equation}
l(f_{\theta}, \mathcal{M}_{k}) = \frac{1}{|\mathcal{M}_{k}|}\sum_{(\mathbf{x}_{i}, k, y_{i}) \in \mathcal{M}_{k}}l(f_{\theta}(\mathbf{x}_{i}, k), y_{i}).
\end{equation}

GEM avoids catastrophic forgetting by storing an episodic memory $\mathcal{M}_{k}$ for each task $k$; therefore, the intuitive idea is that it guarantees to avoid increasing the losses over episodic memories of tasks. % to avoid
%Editor: On the line below, please ensure that the intended meaning has been maintained in this edit.
%increasing the loss.
Formally, for task $t$, GEM solves for the following objective:
\begin{equation}
\begin{array}{ll}
\text{minimize}_{\theta} & l(f_{\theta}, \mathcal{D}_{t})  \\
\text{subject to}  & l(f_{\theta}, \mathcal{M}_{k}) - l(f_{\theta}^{t-1}, \mathcal{M}_{k}) \le 0, \quad \forall k < t, \label{s_gem_eq_1}
\end{array}
\end{equation}
where $f_{\theta}^{t-1}$ is the trained network until task $t-1$. The dual optimization problem of GEM is given by:

\begin{equation}
\begin{array}{ll}
\text{minimize}_{\widetilde{g}} & \frac{1}{2} || g - \widetilde{g} ||_{2}^{2} \\
\text{subject to} &  \langle \widetilde{g}, g_{k} \rangle \ge 0, \quad \forall k < t,
\label{s_gem_eq}
\end{array}
\end{equation}
where $\langle g, g_{k} \rangle = \left\langle \frac{\partial l(f_{\theta}(\mathbf{x},t), y)}{\partial \theta}, \frac{\partial l(f_{\theta}, \mathcal{M}_{k})}{\partial \theta}\right\rangle$.

\begin{figure}[!t]
	\centering
	\subfloat[GEM]{\includegraphics[width=0.48\linewidth]{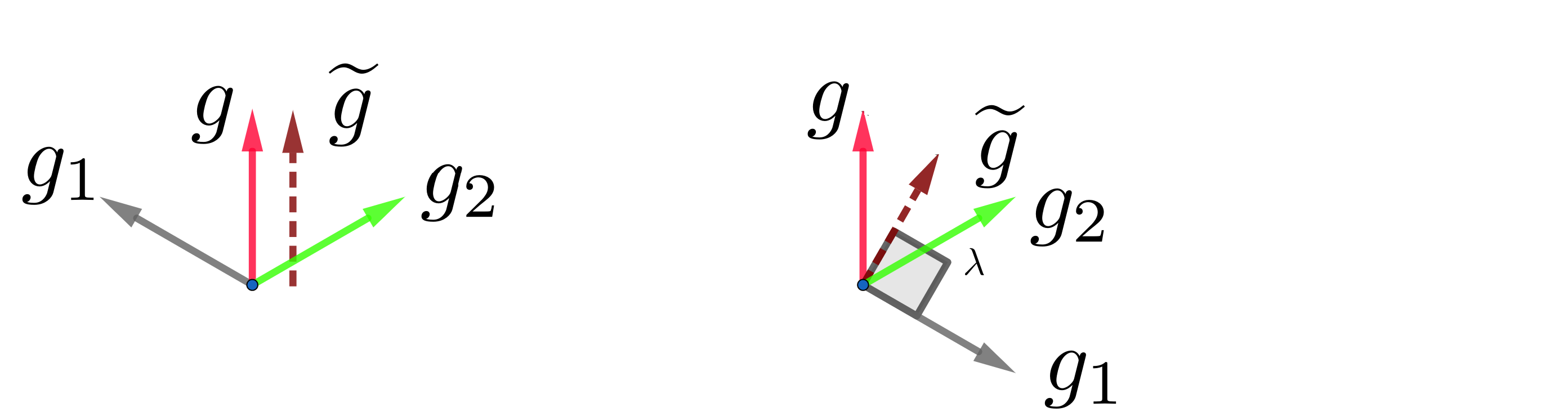}}\quad
	\subfloat[A-GEM]{\includegraphics[width=0.48\linewidth]{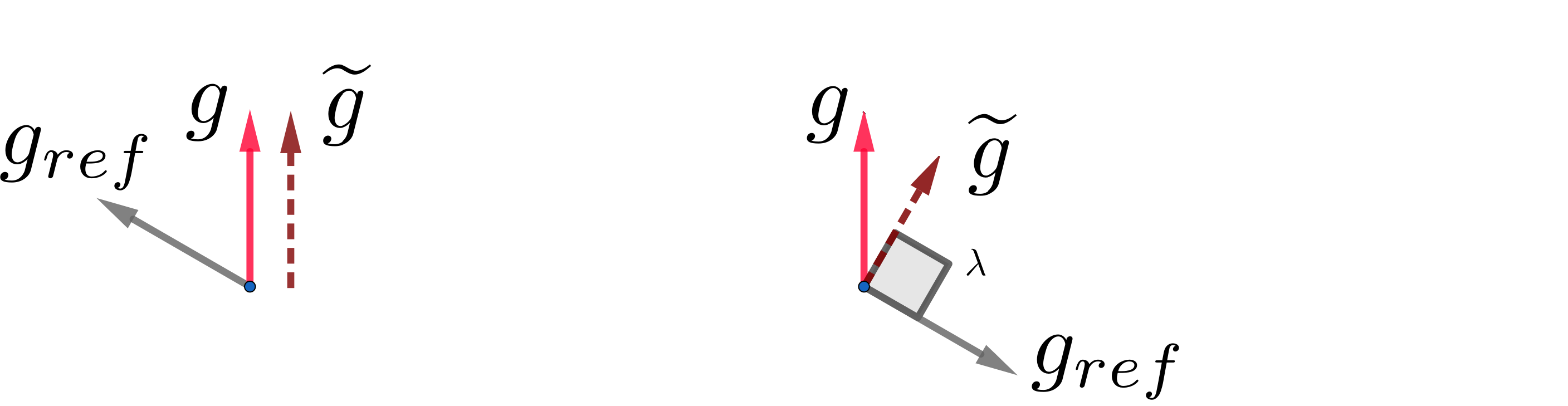}}\\
	\subfloat[$\epsilon$-SOFT-GEM]{\includegraphics[width=0.48\linewidth]{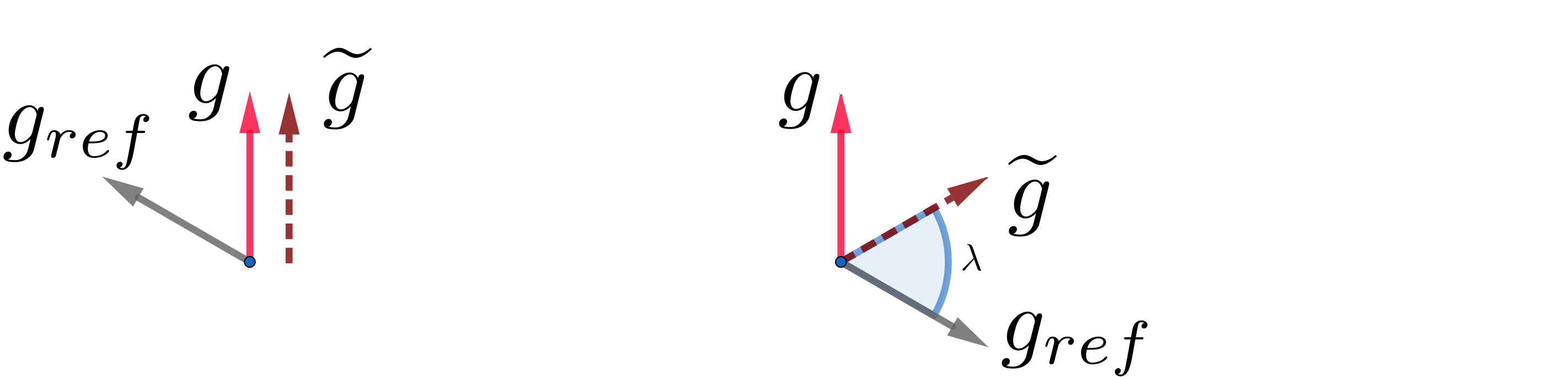}}\quad
	\subfloat[A-A-GEM]{\includegraphics[width=0.48\linewidth]{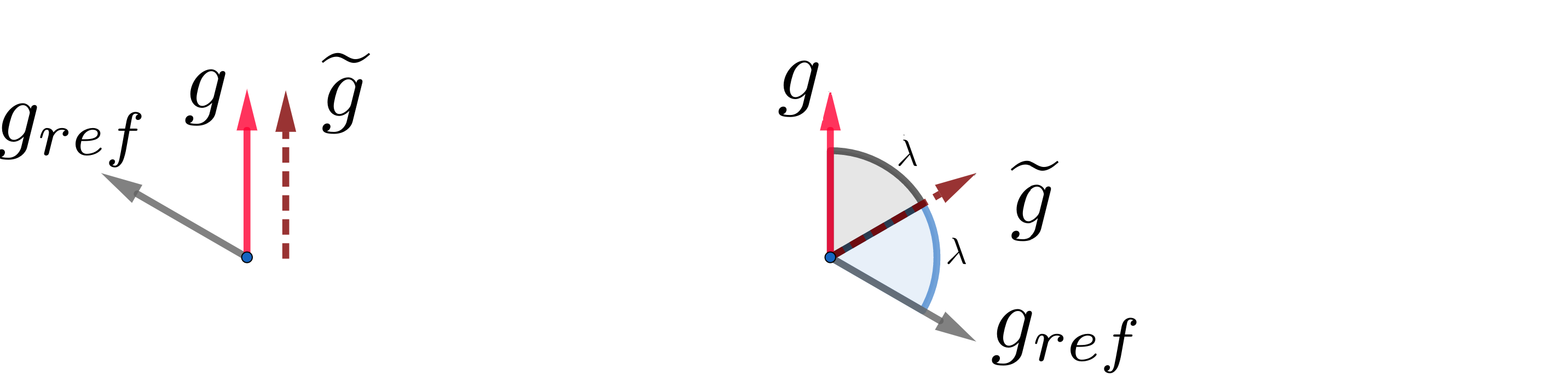}}\\
	\caption{The basic ideas of GEM, A-GEM, $\epsilon$-SOFT-GEM and A-A-GEM.}
	\label{fig:fig_gems_idea}
\end{figure}

The basic idea of GEM is shown in \textbf{Figure} \ref{fig:fig_gems_idea}(a), where $g_{1}$ and $g_{2}$ are the gradients of samples from $\mathcal{M}_{1}$ and $\mathcal{M}_{2}$ respectively, $g$ is the gradient of samples from the current task, $\widetilde{g}$ is the updated gradient, and $\lambda=90^{\circ}$.

\subsection{A-GEM}
GEM computes the matrix $G=-(g_1, g_2, ..., g_{t-1})$ by using samples from episodic memory, which requires much computation.
The average GEM (A-GEM) \cite{Chaudhry2018EfficientLL} aims to ensure that at every training step, the average episodic memory loss over the previous tasks does not increase. Formally, the objective of A-GEM is:
\begin{equation}
\begin{array}{ll}
\text{minimize}_{\theta} & l(f_{\theta}, \mathcal{D}_{t}) \\
\text{subject to} & l(f_{\theta}, \mathcal{M}) \le l(f_{\theta}^{t-1}, \mathcal{M}), \mathcal{M} = \bigcup_{k<t}\mathcal{M}_{k}, \label{s_gem_eq_10}
\end{array}
\end{equation}
and the dual optimization problem is:
\begin{equation}
\begin{array}{ll}
\text{minimize}_{\widetilde{g}} & \frac{1}{2} || g - \widetilde{g} ||_{2}^{2}  \\
\text{subject to} & \langle \widetilde{g}, g_{ref} \rangle  \ge 0, \label{s_agem_eq}
\end{array}
\end{equation}
where $(\mathbf{x}_{ref}, y_{ref}) \sim \mathcal{M}$ and $g_{ref}$ is a gradient on a batch of $(\mathbf{x}_{ref}, y_{ref})$.

The update rule of A-GEM is:
\begin{equation}
\begin{array}{l}
\widetilde{g} = g - \frac{{g}^{\mathrm{T}}g_{ref}}{g_{ref}^{\mathrm{T}}g_{ref}}g_{ref}.
\end{array}
\end{equation}

The basic idea is shown in \textbf{Figure} \ref{fig:fig_gems_idea}(b), where $\lambda =90^{\circ}$ and $\cos \lambda = \cos 90^{\circ} = 0$.
\begin{equation}
\begin{array}{l}
g_{ref}^{\mathrm{T}}\widetilde{g} = g_{ref}^{\mathrm{T}}(g - \frac{{g}^{\mathrm{T}}g_{ref}}{g_{ref}^{\mathrm{T}}g_{ref}}g_{ref}) = 0
\end{array}
\end{equation}

\subsection{$\epsilon$-SOFT-GEM}

The basic ideas shown in Figure \ref{fig:fig_gems_idea}(a) and \ref{fig:fig_gems_idea}(b) and the update rules of GEM and A-GEM show that the loss over episodic memory for previous tasks is unchanged. However, we believe that the model can decrease the loss over episodic memory while learning a new task. The optimization problem of $\epsilon$-SOFT-GEM is given by:
\begin{equation}
\begin{array}{ll}
\text{minimize}_{\widetilde{g}} & \frac{1}{2} || g - \widetilde{g} ||_{2}^{2} \\
\text{subject to} & \langle \widetilde{g}, g_{ref} \rangle \ge \epsilon, \epsilon \in [0, 1],  \label{eq_soft_gem_1}
\end{array}
\end{equation}
where $\epsilon$ is a soft constraint that balances the capacities for learning new information and preserving the old learned information.

We introduce $\hat{g}$ as a normalized gradient of $g$ and $\hat{g}_{ref}$ as a normalized gradient of $g_{ref}$, where $\hat{g} = \frac{g}{|g|}$ and $\hat{g}_{ref} = \frac{g_{ref}}{|g_{ref}|}$.
Rewriting (\ref{eq_soft_gem_1}) yields:
\begin{equation}
\begin{array}{ll}
\text{minimize}_{\widetilde{g}} & \frac{1}{2} || \hat{g} - \widetilde{g} ||_{2}^{2} \\
\text{subject to} & \widetilde{g}^{\mathrm{T}} \hat{g}_{ref}  \ge \epsilon, \epsilon \in [0, 1]. \label{eq_soft_gem_2}
\end{array}
\end{equation}

From the basic idea of $\epsilon$-SOFT-GEM shown in \textbf{Figure} \ref{fig:fig_gems_idea}(c), we can conclude that the loss over episodic memory and new tasks can both decrease at the update step.

The update rule of $\epsilon$-SOFT-GEM is:
\begin{equation}
\begin{array}{l}
\widetilde{g} = \hat{g} - \frac{{\hat{g}}^{\mathrm{T}}\hat{g}_{ref} - \epsilon}{\hat{g}_{ref}^{\mathrm{T}}\hat{g}_{ref}}\hat{g}_{ref},
\end{array}
\end{equation}
we can conclude that $\cos \lambda = \epsilon$.

The formal proof of the update rule of $\epsilon$-SOFT-GEM is given in \textbf{Appendix} \ref{SOFT_GEM_UPDATE_RULE}, and the algorithm is described as \textbf{Algorithm} \ref{alg:soft_gem} in \textbf{Appendix} \ref{appendix_soft_gem_alg}.

\subsection{Average A-GEM (A-A-GEM)}
In this section, we introduce an intuitive idea to decrease the loss over old tasks and novel tasks. The basic idea of average A-GEM (A-A-GEM) is shown in \textbf{Figure} \ref{fig:fig_gems_idea}(d).

The update rule of A-A-GEM is:
\begin{equation}
\begin{array}{l}
\widetilde{g} = \frac{1}{2}(\hat{g} + \hat{g}_{ref})
\end{array}
\label{s_a_agem_eq}
\end{equation}
the update rule can ensure that $\lambda \le 90^{\circ}$.

The algorithm is described as \textbf{Algorithm} \ref{alg:soft_gem} in \textbf{Appendix} \ref{appendix_soft_gem_alg}.

\section{Experiment}
\subsection{Sequential tasks}
In this section, we consider 4 data streams, which can be classified into 3 sequential tasks:
\begin{enumerate}
	\item \textbf{Independent sequential tasks}. Independent sequential tasks have no information to share; \textbf{Permuted MNIST} \cite{Kirkpatrick2017OvercomingCF} is a standard sequential task that is a variant of the MNIST \cite{726791} handwritten digit database where the pixels of images are shuffled by a fixed random permutation sequence in each task.
	\item \textbf{Sequential tasks with new classes}. In learning sequential tasks for new classes, we assume the model trains on disjoint data sequentially. \textbf{Split CIFAR} \cite{Zenke2017ContinualLT} splits the original CIFAR-100 \cite{Krizhevsky2009LearningML} dataset into 20 disjoint subsets, where
	%Editor: On the line below, please ensure that the intended meaning has been maintained in this edit.
	every fifth class is randomly sampled
	from 100 classes without overlapping. Similar to Split CIFAR, \textbf{Split CUB} splits the fine-grained image classification dataset CUB \cite{Wah2011TheCB} into 20 disjoint subsets and has a total of 200 bird categories.
	\item \textbf{Sequential tasks sharing some of the classes}. \textbf{Split AWA} is an incremental version of the AWA dataset \cite{Lampert2009Learning} of 50 animal categories, where each task is constructed by sampling 5 classes with replacement from the 50 classes to construct 20 tasks.
\end{enumerate}

A summary of the tasks described above is shown in \textbf{Table} {\ref{dataset_statistics_tab}} in \textbf{Appendix} \ref{data_static}.

Similar to the setting of experiments in \cite{Chaudhry2018EfficientLL}, when training on the Permuted MNIST, Split CIFAR, Split CUB and Split AWA, we cross-validate on the first 3 tasks and then evaluate the metrics on the remaining 17 tasks in a single training epoch over each task in sequence, which means that $T^{CV}=3$ and $T = 20$.

\begin{figure}[htbp]
	\centering
	\hspace*{-1cm}
	\subfloat[Permuted MNIST]{\label{fig:fig_methods_avg_acc_forget_permuted_mnist}\includegraphics[width=0.7\linewidth]{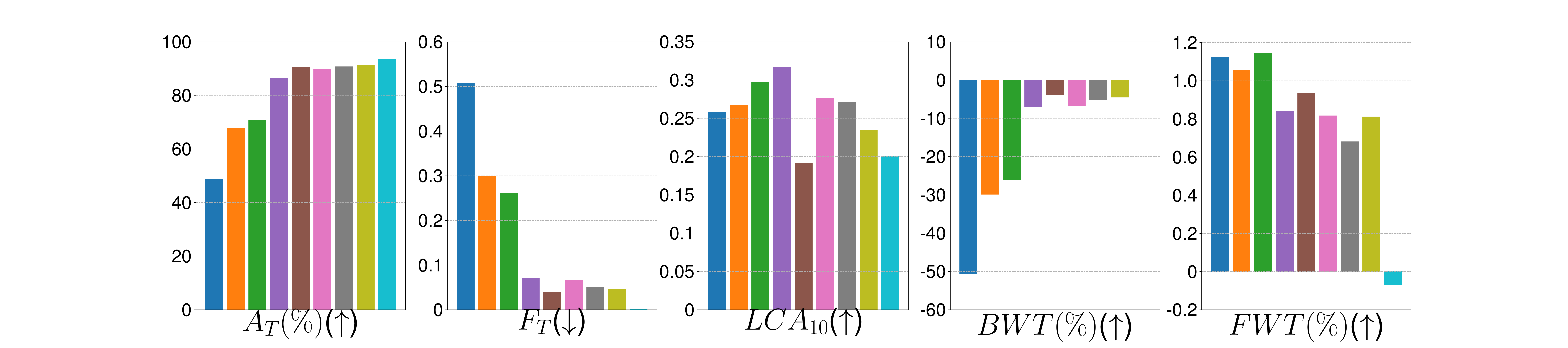}} \\
	\hspace*{-1cm}
	\subfloat[Split CIFAR]{\label{fig:fig_methods_avg_acc_forget_split_cifar}\includegraphics[width=0.7\linewidth]{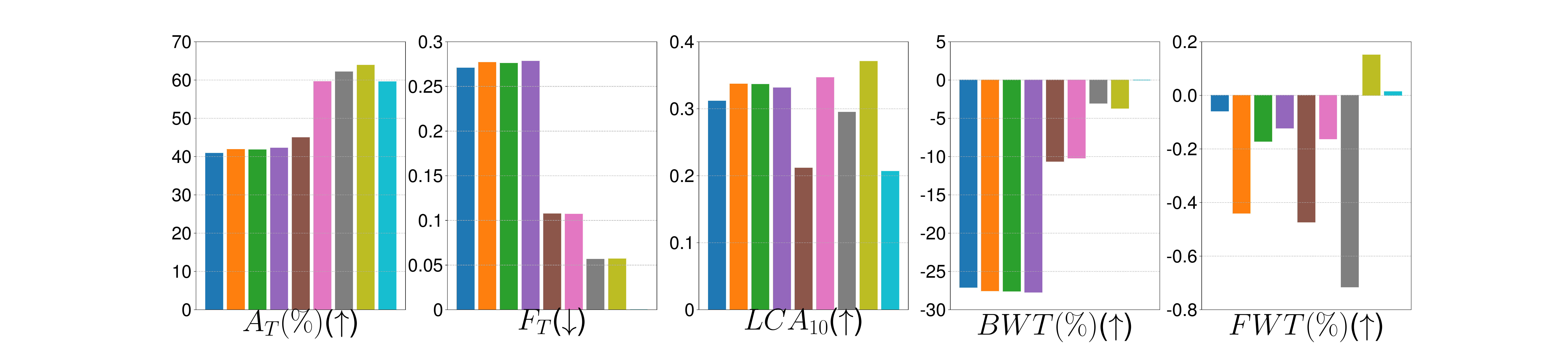}} \\
	\hspace*{-1cm}
	\subfloat[Split CUB]{\label{fig:fig_methods_avg_acc_forget_split_cub}\includegraphics[width=0.7\linewidth]{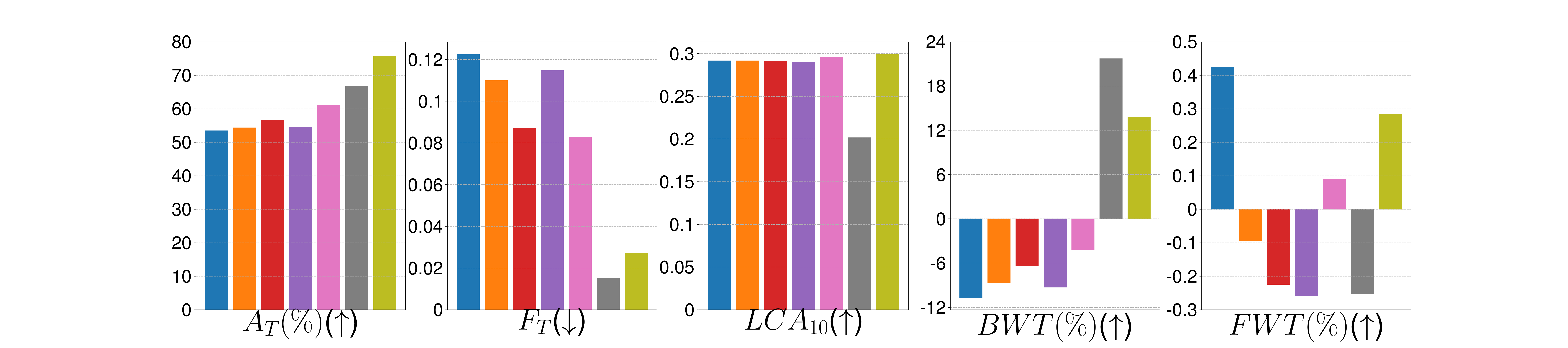}} \\
	\hspace*{-1cm}
	\subfloat[Split CUB(-JE)]{\label{fig:fig_methods_avg_acc_forget_split_cub_je}\includegraphics[width=0.7\linewidth]{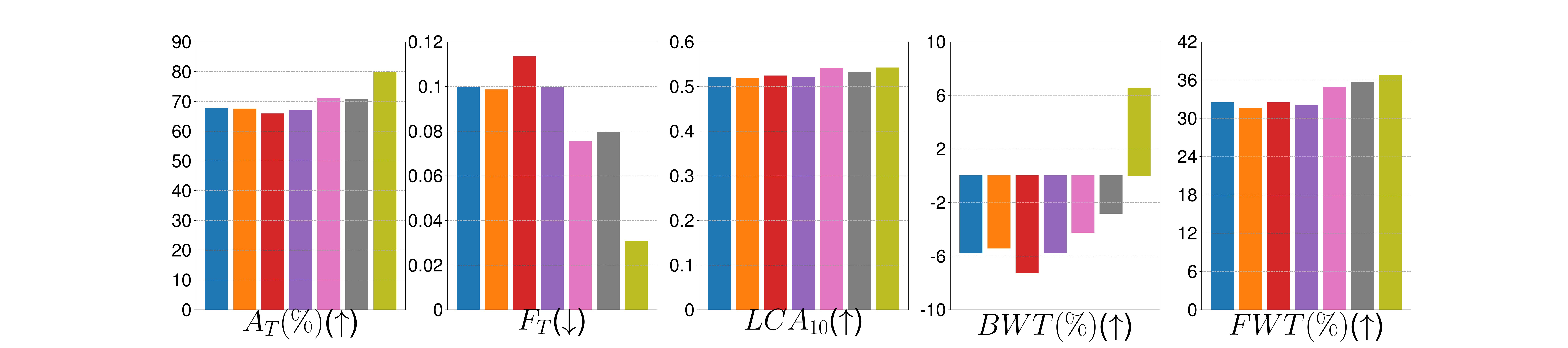}} \\
	\hspace*{-1cm}
	\subfloat[Split AWA]{\label{fig:fig_methods_avg_acc_forget_split_awa}\includegraphics[width=0.7\linewidth]{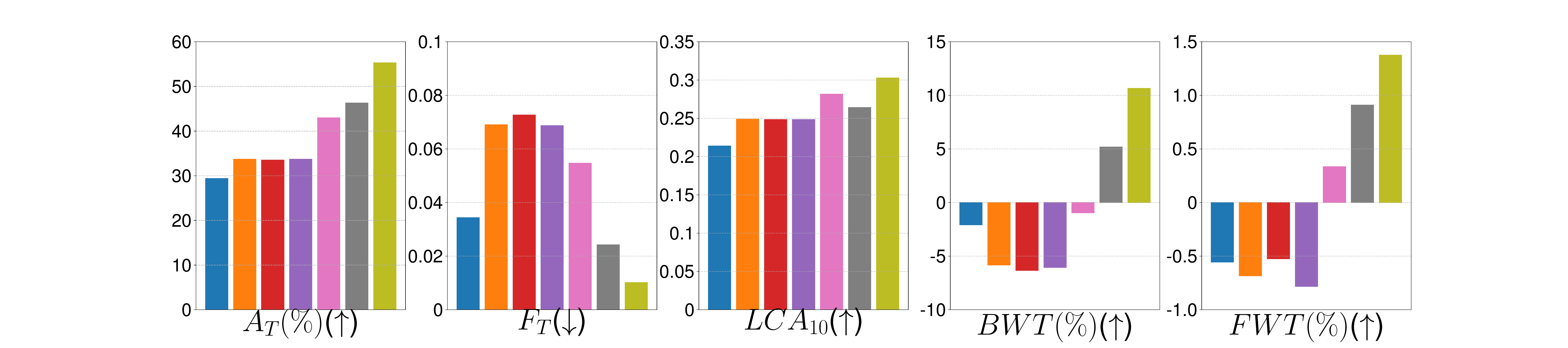}} \\
	\hspace*{-1cm}
	\subfloat[Split AWA(-JE)]{\label{fig:fig_methods_avg_acc_forget_split_awa_je}\includegraphics[width=0.7\linewidth]{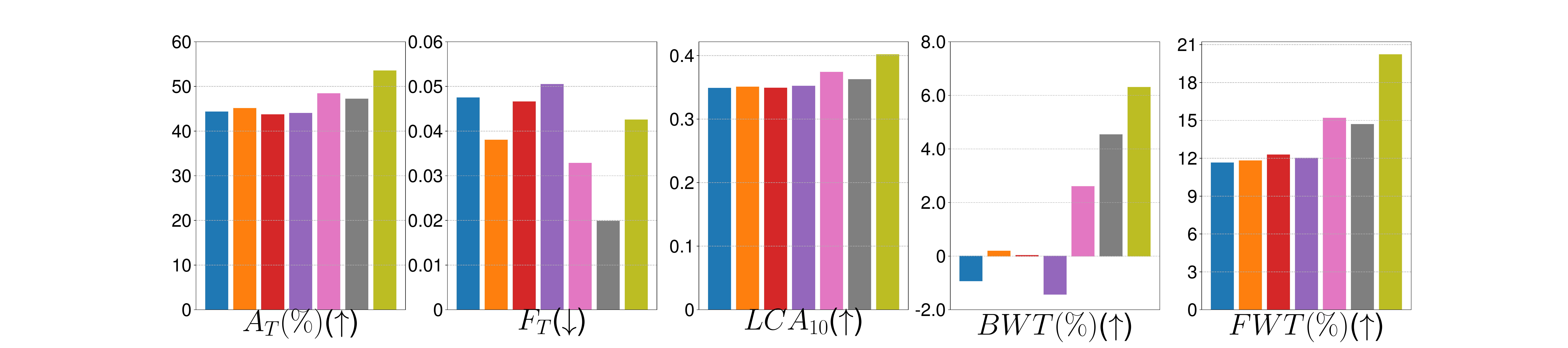}} \\
	
	\subfloat{\label{fig:fig_methods_avg_acc_forget_legend}\includegraphics[width=0.7\linewidth]{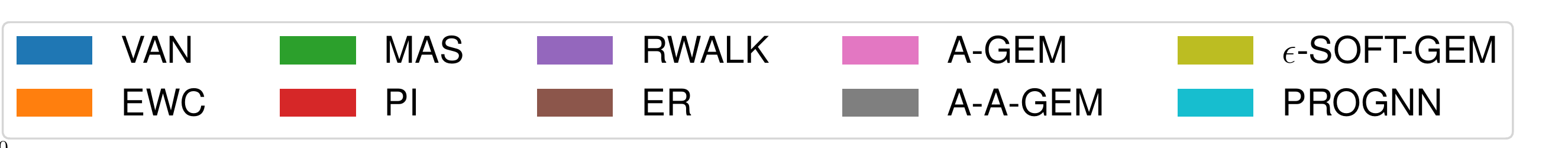}}
	
	\caption{Models evaluated on Permuted MNIST, Split CIFAR, Split CUB, Split CUB(-JE), Split AWA and Split AWA(-JE); the models are trained over 5 runs.}
	
	\label{fig:fig_methods_avg_acc_forget_permuted_mnist_split_cifar_split_cub_split_cub_je}
\end{figure}

\subsection{Baselines}
The baselines and state-of-the-art approaches can be classified into
(\romannumeral1) training the model without regularization or extra memory, where the parameters of the current task are initialized from the parameters of the previous task; VAN is shown in the experiments;
(\romannumeral2) training individual models on previous tasks and then carrying out a new stage of training a new task with models of previous tasks such as PROGNN \cite{Rusu2016ProgressiveNN};
(\romannumeral3) using a regularization to slow down learning on network weights that correlate with previous acquired knowledge, such as EWC \cite{Kirkpatrick2017OvercomingCF}, PI \cite{Zenke2017ContinualLT}, RWALK \cite{Chaudhry2018RiemannianWF} and MAS \cite{Aljundi2018MemoryAS};
(\romannumeral4) using extra memory to provide the data for previous tasks in a sustained manner. A-GEM, A-A-GEM and $\epsilon$-SOFT-GEM are combinations of regularization and extra memory.
In addition, a generative model \cite{Shin2017ContinualLW,Kamra2018Deep} used as extra memory is applied in continual learning as well; it is not considered in this work because it performs poorly in a single training epoch.
Finally, we refer to one-hot embedding as the default, and to joint embedding \cite{Chaudhry2018EfficientLL} by appending a suffix '-JE' to the sequential task name, which indicates that the models are trained and evaluated on the task by adopting joint embedding.

The settings of the neural network architectures used in this paper are described in \textbf{Table} \ref{tab:setting_architecture_datasets} in \textbf{Appendix} \ref{architecture}.
For a given sequential task, all models use the same architecture and apply stochastic gradient descent optimization with 10 for the mini-batch size;
the remaining training parameters of the models are the same as in \cite{Chaudhry2018EfficientLL}.

\subsection{Metrics}
We evaluated the model on the following metrics:
\begin{enumerate}
	\item The average accuracy on all tasks after the $k$-th sequential task learned $A_{k}$ \cite{LopezPaz2017GradientEM}, which indicates the balance of the stability and plasticity of the model; $A_{T}$ is the average accuracy of all tasks after the last task learned; the accuracy of the first task after all sequential tasks learned $a_{1}$ and the accuracy of the last task after all sequential tasks learned $a_{t}$ are also considered.
	\item The forgetting $F_{k}$ \cite{Chaudhry2018RiemannianWF}, which indicates the ability of the model to preserve the knowledge of previous tasks.
	\item The learning curve area ($LCA \in [0,1]$) \cite{Chaudhry2018EfficientLL}, which indicates the ability of the model to learn a new task.
	\item Backward transfer ($BWT$) \cite{LopezPaz2017GradientEM}, which indicates the influence of the performance on previous tasks $k < t$ when the model is learning task $t$. A positive $BWT$ indicates an increase in the performance of the previous task, and a large negative $BWT$ indicates catastrophic forgetting.
	\item Forward transfer ($FWT$) \cite{LopezPaz2017GradientEM}, which indicates the influence of the performance on future tasks $k > t$ when the model is learning task $t$. A positive $FWT$ shows that the model can perform ``zero-shot'' learning.
\end{enumerate}

\begin{figure}
	\centering
	\subfloat[Permuted MNIST($A_{T}$)]{\label{fig:fig_agem_pbt_split_cifar_avg_acc_mnist}\includegraphics[width=0.48\linewidth]{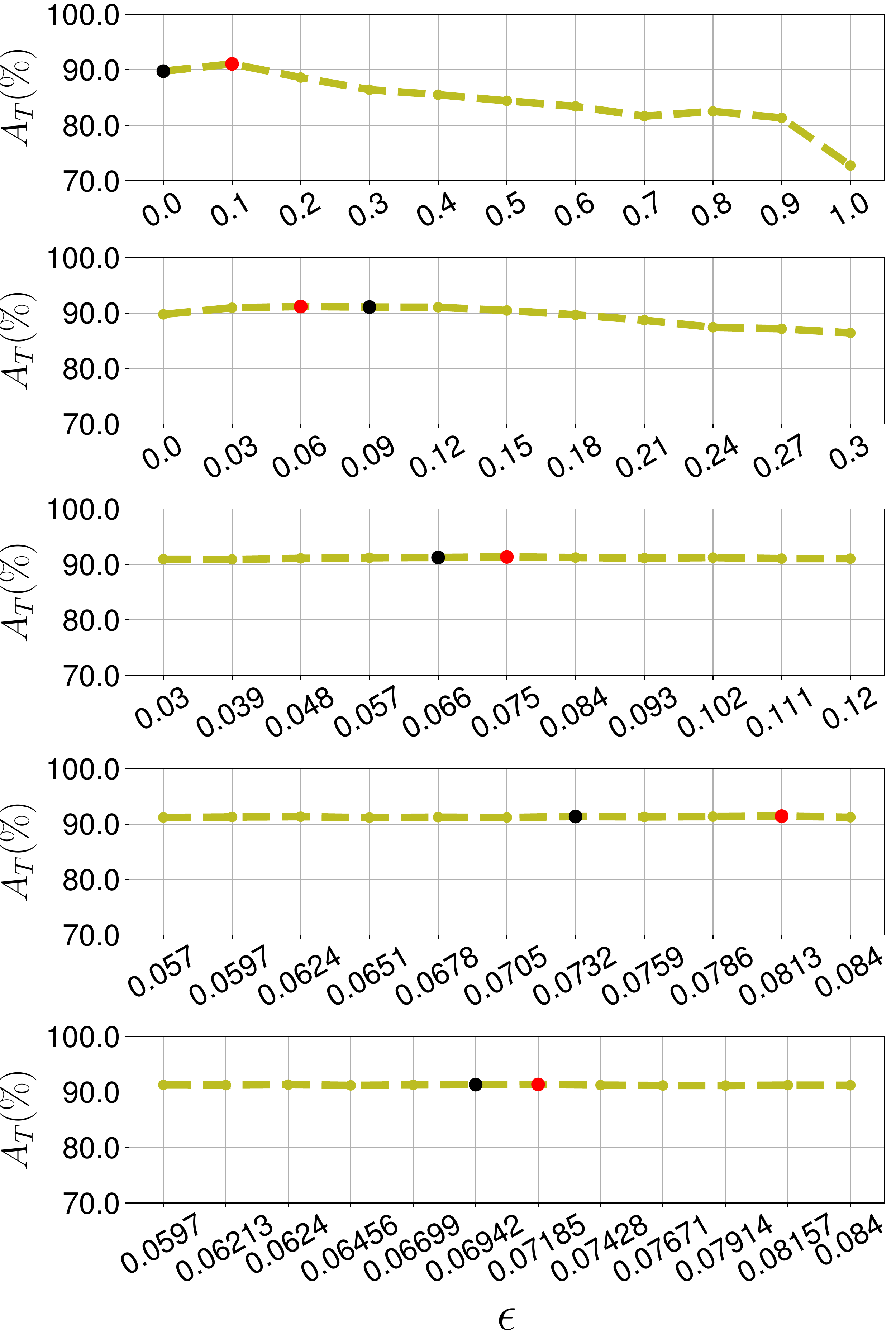}}
	\subfloat[Permuted MNIST($F_{T}$)]{\label{fig:fig_agem_pbt_split_cifar_fgt_mnist}\includegraphics[width=0.48\linewidth]{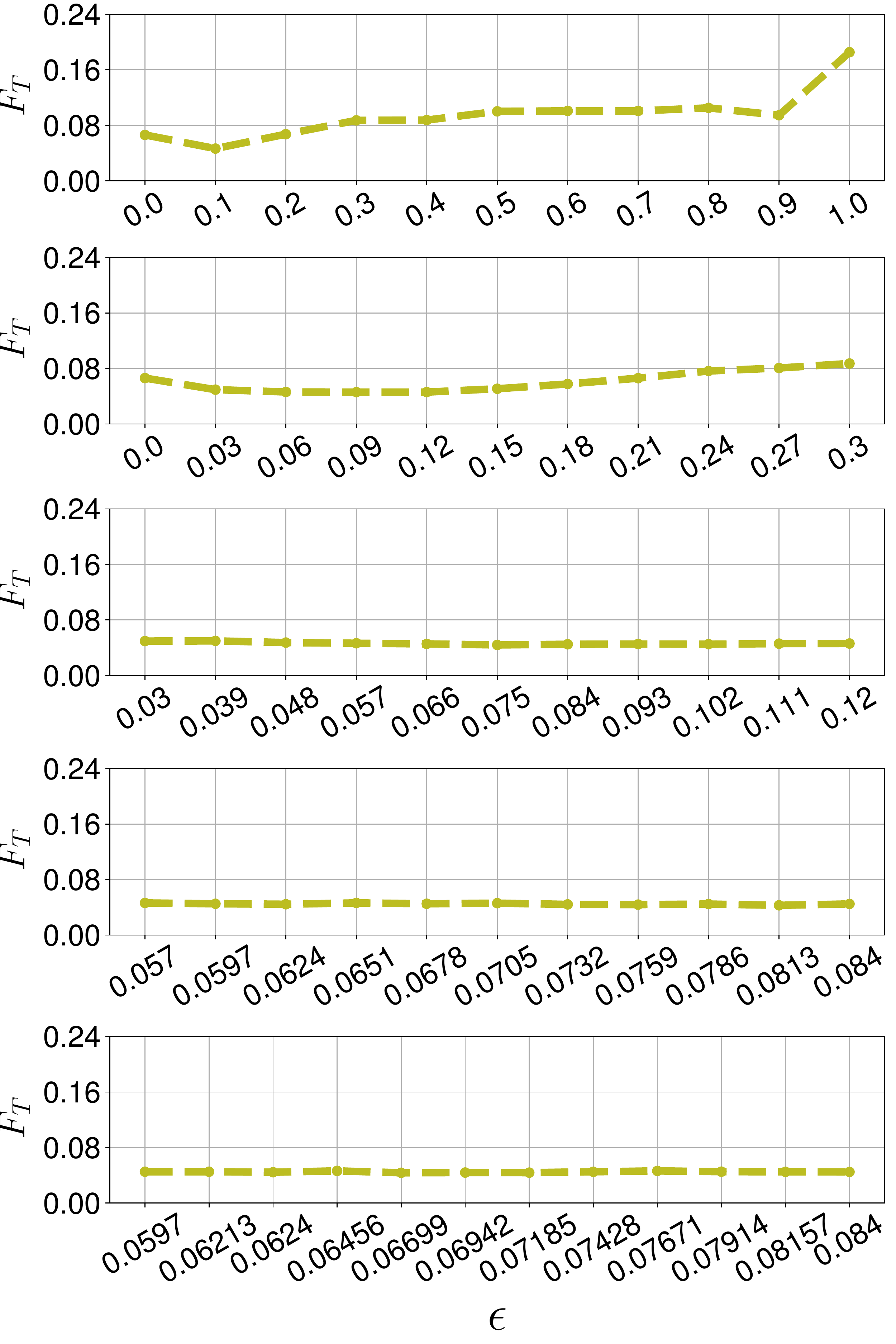}} \\
	\caption{$\epsilon$-SOFT-GEM on Permuted MNIST in 5 training repeats, where the models are trained over 5 runs in a single training repeat.}
	\label{fig:fig_agem_pbt_permuted_mnist_split_cifar_avg_acc_fgt_mnist}
\end{figure}

\begin{figure}
	\centering
	\subfloat[Split CIFAR($A_{T}$)]{\label{fig:fig_agem_pbt_split_cifar_avg_acc_cifar}\includegraphics[width=0.48\linewidth]{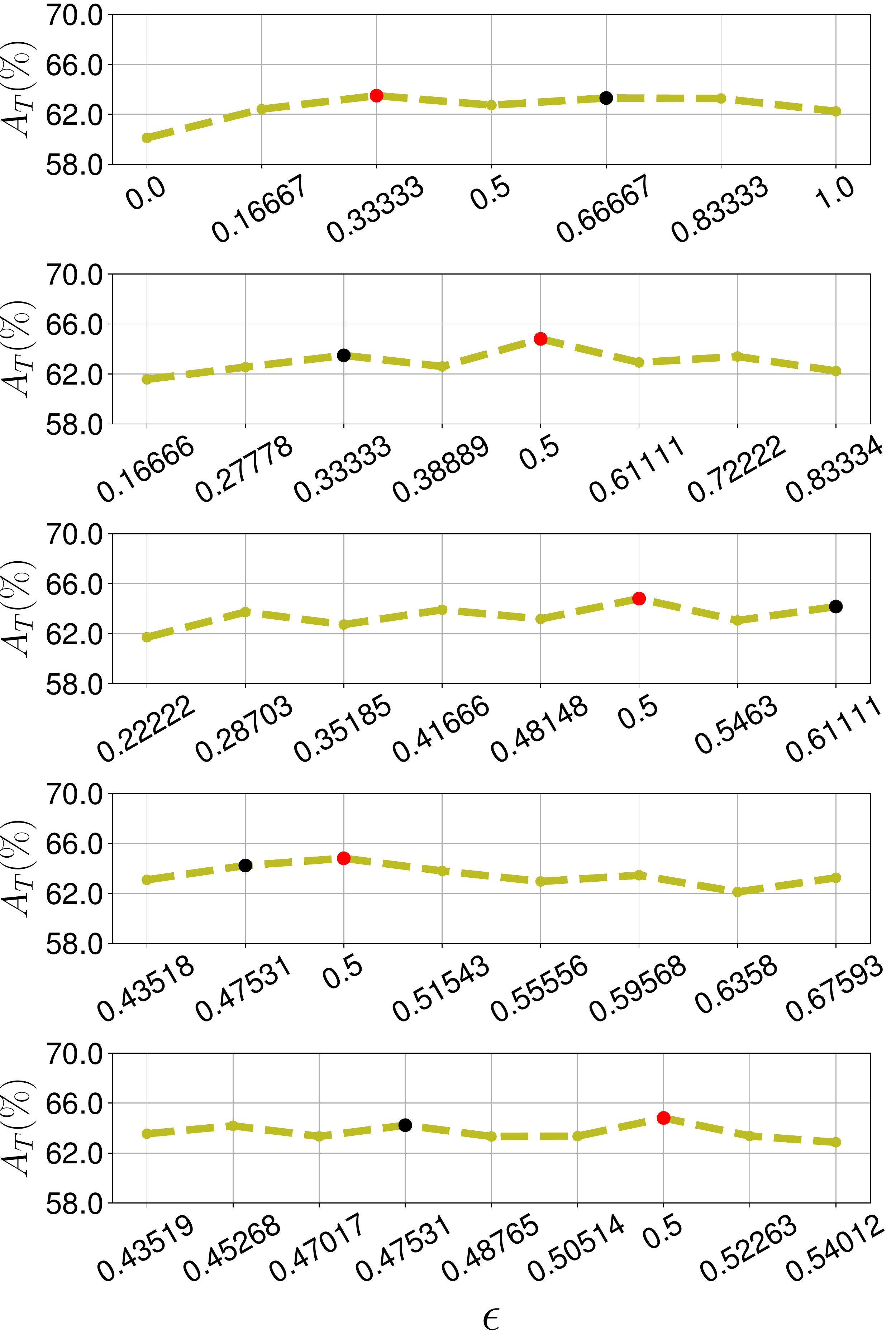}}
	\subfloat[Split CIFAR($F_{T}$)]{\label{fig:fig_agem_pbt_split_cifar_fgt_cifar}\includegraphics[width=0.48\linewidth]{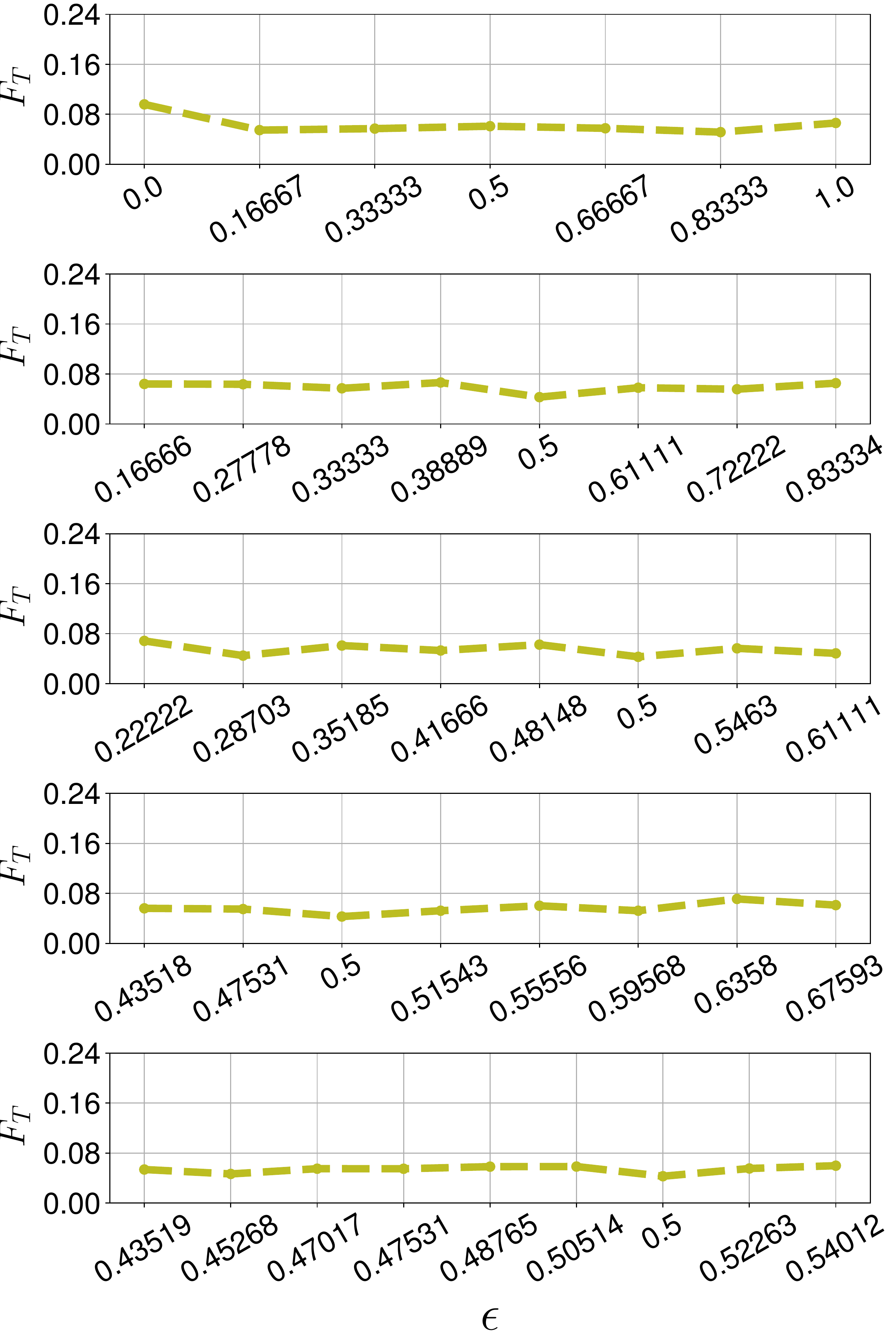}}
	\caption{$\epsilon$-SOFT-GEM on Split CIFAR in 5 training repeats, where the models are trained over 5 runs in a single training repeat.}
	\label{fig:fig_agem_pbt_permuted_mnist_split_cifar_avg_acc_fgt_cifar}
\end{figure}

\subsection{Comparison with baselines}
In this section, we show the applicability of $\epsilon$-SOFT-GEM and A-A-GEM on sequential learning tasks. The details of the results are shown in \textbf{Tables} \ref{tab:dataset_mnist_cifar_statistics_permuted_mnist}, \ref{tab:dataset_mnist_cifar_statistics_cifar}, \ref{tab:dataset_cub_statistics_tab_ohot}, \ref{tab:dataset_cub_statistics_tab_je}, \ref{tab:dataset_awa_statistics_tab_ohot} and \ref{tab:dataset_awa_statistics_tab_je} in \textbf{Appendix} \ref{RESULTS}.

First, \textbf{Figure} \ref{fig:fig_methods_avg_acc_forget_permuted_mnist_split_cifar_split_cub_split_cub_je}
shows that $\epsilon$-SOFT-GEM outperforms other models on Permuted MNIST, Split CIFAR, Split CUB, Split CUB(-JE), Split AWA and Split AWA(-JE), except for PROGNN, which achieves slightly better performance than $\epsilon$-SOFT-GEM on Permuted MNIST.
The reason is that PROGNN trains an individual model on a previous task and then carries out a new stage of training a new task; it can preserve all the information it learned on previous tasks.
Meanwhile, from the snapshot of the statistics of the datasets shown in \textbf{Table} \ref{dataset_statistics_tab} in \textbf{Appendix} \ref{data_static}, PROGNN achieves better performance on a large-scale training sample dataset and lower performance on a smaller training sample dataset.
However, PROGNN has the worst memory problem because the size of the parameters of the model increases superlinearly with the number of tasks, and it will run out of memory during training due to the large size of the model; therefore, PROGNN is invalid on Split CUB and Split AWA, which apply the standard ResNet18 in \textbf{Table} \ref{tab:setting_architecture_datasets} in \textbf{Appendix} \ref{architecture}, which is not shown in \textbf{Figures}
\ref{fig:fig_methods_avg_acc_forget_split_cub}, \ref{fig:fig_methods_avg_acc_forget_split_cub_je}, \ref{fig:fig_methods_avg_acc_forget_split_awa} and \ref{fig:fig_methods_avg_acc_forget_split_awa_je}.

Second, $\epsilon$-SOFT-GEM acquires better $a_{1}$, $a_{t}$, $A_{T}$ and $F_{T}$ values than the other baselines shown in \textbf{Tables} \ref{tab:dataset_mnist_cifar_statistics_permuted_mnist}, \ref{tab:dataset_mnist_cifar_statistics_cifar}, \ref{tab:dataset_cub_statistics_tab_ohot}, \ref{tab:dataset_cub_statistics_tab_je}, \ref{tab:dataset_awa_statistics_tab_ohot} and \ref{tab:dataset_awa_statistics_tab_je},
which means that $\epsilon$-SOFT-GEM can maintain its performance on previous tasks when learning new tasks.
$LCA_{10}$ in \textbf{Figure} \ref{fig:fig_methods_avg_acc_forget_permuted_mnist_split_cifar_split_cub_split_cub_je}
shows that $\epsilon$-SOFT-GEM has a competitive capacity to learn new knowledge fast, even compared with A-GEM.

Third, the $BWT$ values of $\epsilon$-SOFT-GEM and A-A-GEM are positive in Split CUB and Split AWA; \textbf{Figure} \ref{fig:fig_methods_avg_acc_forget_permuted_mnist_split_cifar_split_cub_split_cub_je} shows that $\epsilon$-SOFT-GEM and A-A-GEM can learn new knowledge of previous tasks to increase the performance of the model on previous tasks, while the other baselines have a negative $BWT$ for 4 sequential learning tasks. The $FWT$ value indicates that $\epsilon$-SOFT-GEM has a competitive ability to perform ``zero-shot'' learning.

Fourth, from the results shown in \textbf{Figure} \ref{fig:fig_methods_avg_acc_forget_permuted_mnist_split_cifar_split_cub_split_cub_je},
A-A-GEM performs better than the other models except for $\epsilon$-SOFT-GEM, but A-A-GEM is much simpler than $\epsilon$-SOFT-GEM without specifying $\epsilon$.

Finally, we can conclude that $\epsilon$-SOFT-GEM can balance preserving the knowledge of old tasks with a soft constraint $\epsilon$ and learning new tasks with a fast learning curve.

\subsection{Exploration of $\epsilon$}

In this work, we use a simple heuristic algorithm to explore $\epsilon$. Each row in \textbf{Figures}
\ref{fig:fig_agem_pbt_permuted_mnist_split_cifar_avg_acc_fgt_mnist} and \ref{fig:fig_agem_pbt_permuted_mnist_split_cifar_avg_acc_fgt_cifar}
is a population trained with a specific set $\epsilon$, and there are 5 training repeats in the
%Editor: On the line below, please ensure that the intended meaning has been maintained in this edit.
whole experiment.

In the first repeat, we divide $\epsilon \in [0, 1]$ into $N$ equal parts with an interval $\delta$; for example, for Permuted MNIST, $N=11, \delta=0.1$, and for Split CIFAR, $N=7, \delta=0.16667$.

After the $j$-th training repeat, we choose the $\epsilon_{j,1}$ that yields the best $A_{j,T}$, the $\epsilon_{j,2}$ with the second-best $A_{j,T}$ and an interval $\delta_{j}$, where $A_{j, T}$ is $A_{T}$ after the $j$-th training repeat and $\delta_{j}$ is the interval of the $j$-th training repeat.
Meanwhile, we define $\epsilon_{j}[1]$ and $\epsilon_{j}[N]$ as the smallest and largest values of $\epsilon_{j}$, respectively. We assume $\epsilon_{j, 1} \le \epsilon_{j, 2}$, and the update rule of $\epsilon_{j+1}$ is:
\setlength{\arraycolsep}{0.0em}
\begin{equation}
\footnotesize
\epsilon_{j+1} \in
\left\{
\begin{array}{l}
\left[0, 1\right], \quad j=0; \\
\mathrm{stop}, \quad \epsilon_{j, 1}=\epsilon_{j}[1], \epsilon_{j, 2} = \epsilon_{j}[N]  \  \mathrm{and} \  j > 0 \ \mathrm{or} \ j > M\\
\left[\epsilon_{j,1}, \epsilon_{j,2}+\delta_{j}\right], \quad \epsilon_{j,1}=\epsilon_{j}[1] \  \mathrm{and} \  0 < j \le M;  \\
\left[\epsilon_{j, 1} - \delta_{j}, \epsilon_{j,2}\right], \quad \epsilon_{j,2} = \epsilon_{j}[N] \  \mathrm{and} \  0 < j \le M; \\
\left[\epsilon_{j, 1} - \delta_{j}, \epsilon_{j, 2} + \delta_{j}\right], \  \mathrm{other};
\end{array}
\right.
\end{equation}
where $M$ is the number of training repeats. $\epsilon$-SOFT-GEM with $\epsilon=0.0$ is equal to A-GEM with the original $g$ and $g_{ref}$, not $\hat{g}$ and $\hat{g}_{ref}$.

The soft constraint $\epsilon$ adjusts the capacity of the model to learn new tasks and preserve old tasks.
According to the update rule above, we repeat the process 5 times to explore $\epsilon$.
First, the top row in \textbf{Figures} \ref{fig:fig_agem_pbt_permuted_mnist_split_cifar_avg_acc_fgt_mnist} and \ref{fig:fig_agem_pbt_permuted_mnist_split_cifar_avg_acc_fgt_cifar}
shows that $\epsilon$-SOFT-GEM outperforms A-GEM with a specified $\epsilon$, such as $\epsilon=0.1$ in Permuted MNIST and all training populations with a specified $\epsilon$ in Split-CIFAR. Second, $A_{T}$ in every repeat
%Editor: On the line below, please ensure that the intended meaning has been maintained in this edit.
is basically a parabolic curve;
therefore, the heuristic optimization algorithm for exploring the best $\epsilon$ is effective. Finally, we find that $\epsilon=0.07185$ in Permuted MNIST and $\epsilon=0.5$ in Split CIFAR after 5 training repeats.

\begin{figure}
	\centering
	\subfloat[Permuted MNIST]{\includegraphics[width=0.48\linewidth]{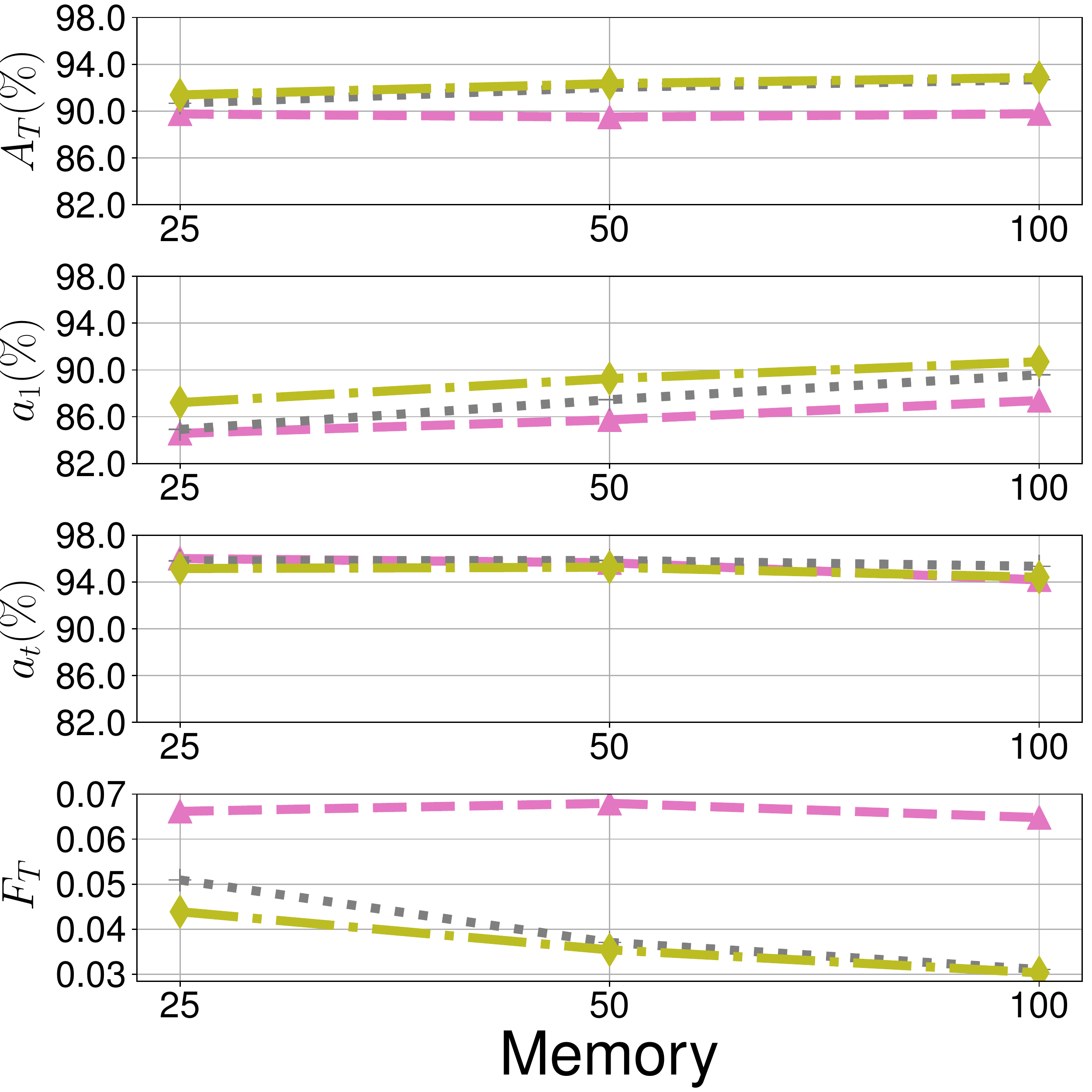}}
	\subfloat[Split CIFAR]{\includegraphics[width=0.48\linewidth]{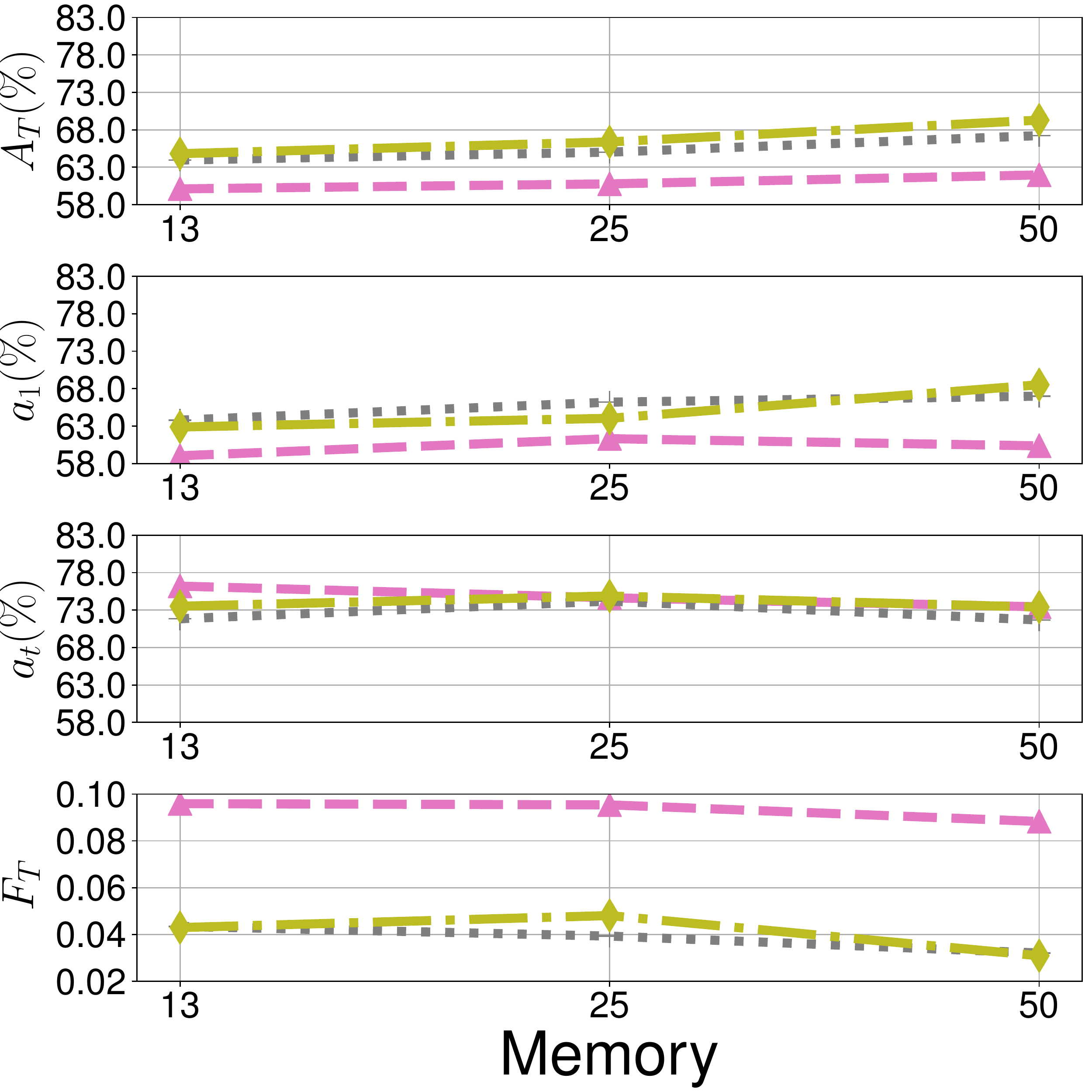}} \\
	\subfloat{\includegraphics[width=0.48\linewidth]{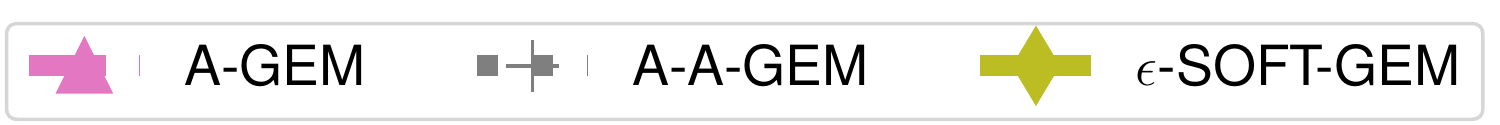}} \\
	\caption{$A_{T}$, $a_{1}$, $a_{t}$ and $F_{T}$ on Permuted MNIST and Split CIFAR with varying episodic memory size in a single training epoch; the models are trained over 5 runs. The details are shown in \textbf{Tables} \ref{tab:dataset_permuted_mnist_episodic_memory_statistics_mnist} and \ref{tab:dataset_permuted_mnist_episodic_memory_statistics_cifar} in \textbf{Appendix} \ref{RESULTS}.}
	\label{fig:fig_avg_acc_mem_permuted_mnist_split_cifar}
\end{figure}

\subsection{Episodic memory}
The conventional solution to catastrophic forgetting is to learn a new task alongside the old samples we
%Editor: On the line below, please ensure that the intended meaning has been maintained in this edit.
preserve; the episodic memory needed to preserve the old samples is significant in A-GEM, A-A-GEM and $\epsilon$-SOFT-GEM.

Therefore, we run the experiments on A-GEM, A-A-GEM and $\epsilon$-SOFT-GEM with varying episodic memory size.
$A_{T}$, $a_{1}$, $a_{t}$ and $F_{T}$ on Permuted MNIST and Split CIFAR are shown in \textbf{Figure} \ref{fig:fig_avg_acc_mem_permuted_mnist_split_cifar}; $\epsilon$-SOFT-GEM outperforms A-GEM and A-A-GEM in $A_{T}$ and $F_{T}$. The reasons are: (\romannumeral1) the larger the episodic memory is, the more old information can be preserved; $g_{ref}$ can represent the actual gradient of the old tasks more accurately, and $\epsilon$-SOFT-GEM can preserve more of the old information, as illustrated in $a_{1}$ which is increasing; (\romannumeral2) A-GEM, A-A-GEM and $\epsilon$-SOFT-GEM can learn new tasks with competitive accuracy in a training epoch with a relatively slow learning curve, which is illustrated by $a_{t}$ which is decreasing.

\subsection{Efficiency}
From the update rule of the gradient, $\epsilon$-SOFT-GEM and A-A-GEM have only one more gradient normalization operation than A-GEM, and we can deduce that $\epsilon$-SOFT-GEM and A-A-GEM acquire almost the same efficient computation and memory costs as A-GEM.

\section{CONCLUSION}
In the real world, humans can learn and accumulate knowledge throughout their whole lives, but ANNs that learn sequential tasks suffer from catastrophic forgetting, in which the learned knowledge is disrupted while a new task is being learned.
To alleviate catastrophic forgetting, we propose a variant of A-GEM with a soft constraint $\epsilon$, called $\epsilon$-SOFT-GEM, as well as A-A-GEM. The experiments demonstrate that $\epsilon$-SOFT-GEM has competitive performance against state-of-the-art models.
First, compared to regularization-based approaches, $\epsilon$-SOFT-GEM achieves significantly higher average accuracy and lower forgetting; additionally, it maintains a fast learning curve and can acquire new knowledge of previous tasks represented by episodic memory when learning new tasks.
Second, $\epsilon$-SOFT-GEM has almost the same efficiency cost as A-GEM in terms of computation and memory. Third, A-A-GEM performs better than the other models except for $\epsilon$-SOFT-GEM, but A-A-GEM is much simpler than $\epsilon$-SOFT-GEM without specifying $\epsilon$.

\section*{Acknowledgement}
The work of this paper is supported by the National Natural Science Foundation of China (No. 91630206), Shanghai Science and Technology Committee (No. 16DZ2293600) and the Program of Shanghai Municipal Education Commission (No. 2019-01-07-00-09-E00018).

\bibliographystyle{unsrt}  

\bibliography{ref}
%\begin{thebibliography}{1}
%
%\bibitem{kour2014real}
%George Kour and Raid Saabne.
%\newblock Real-time segmentation of on-line handwritten arabic script.
%\newblock In {\em Frontiers in Handwriting Recognition (ICFHR), 2014 14th
%  International Conference on}, pages 417--422. IEEE, 2014.
%
%\bibitem{kour2014fast}
%George Kour and Raid Saabne.
%\newblock Fast classification of handwritten on-line arabic characters.
%\newblock In {\em Soft Computing and Pattern Recognition (SoCPaR), 2014 6th
%  International Conference of}, pages 312--318. IEEE, 2014.
%
%\bibitem{hadash2018estimate}
%Guy Hadash, Einat Kermany, Boaz Carmeli, Ofer Lavi, George Kour, and Alon
%  Jacovi.
%\newblock Estimate and replace: A novel approach to integrating deep neural
%  networks with existing applications.
%\newblock {\em arXiv preprint arXiv:1804.09028}, 2018.
%
%\end{thebibliography}

\appendix
%\section*{Appendix A. Probability Distributions for N-Queens}
\section{Sequential Task Statistics}
\label{data_static}
The summary of the sequential tasks described above is shown in \textbf{Table} \ref{dataset_statistics_tab}, where `-' indicates that the size of the samples in each task is different.
\begin{table}[H]
	\small
	\centering
	\caption{Sequential tasks statistics}
	\label{dataset_statistics_tab}
	\begin{tabular}{|l|c|c|c|c|}
		\hline
		& \textbf{Permuted MNIST} & \textbf{Split CIFAR} & \textbf{Split CUB} & \textbf{Split AWA}\\ \hline
		tasks                     & $20$                    & $20$                 & $20$               & $20$\\ \hline
		input size                & $28*28*1$               & $32*32*3$            & $224*224*3$        & $224*224*3$\\ \hline
		classes per task          & $10$                    & $5$                  & $10$               & $5$\\ \hline
		training images per task  & $6*10^4$                & $2.5*10^2$           & $3*10^2$           & - \\ \hline
		test images per task      & $10^4$                  & $5*10^2$             & $2.9*10^2$         & - \\
		\hline
		
	\end{tabular}
\end{table}

\section{Neural Network Architecture}
\label{architecture}

The settings of the neural network architectures used in the paper are described in \textbf{Table} \ref{tab:setting_architecture_datasets}, where memory means the episodic memory per label per task; for example, the memory capacity of each task of Permuted MNIST is $25 \times 10 = 250$, and the total memory capacity is $250 \times 20 = 5000$.
\begin{table}[H]
	\centering
	\caption{The settings of the neural network architectures on sequential tasks}
	\label{tab:setting_architecture_datasets}
	\begin{tabular}{|c|p{180pt}|c|c|}
		\hline
		datasets       & setting                                                         & learning rate & memory   \\ \hline
		Permuted MNIST & Fully-connected network, two hidden layers of 256 ReLU units.                            & 0.1      & 25       \\ \hline
		Split CIFAR    & Reduced ResNet18, same as the model described in \cite{LopezPaz2017GradientEM}. & 0.03     & 13       \\ \hline
		Split CUB      & Standard ResNet18, same as the model described in \cite{He2016DeepRL}.           & 0.1      & 5       \\ \hline
		Split AWA      & Standard ResNet18, same as the model applied on Split CUB. & 0.1 & 20 \\
		\hline
	\end{tabular}
\end{table}

\section{$\epsilon$-Soft-Gem Update Rule}
\label{SOFT_GEM_UPDATE_RULE}
The update rule of $\epsilon$-SOFT-GEM is $\widetilde{g} = \hat{g} - \frac{\hat{g}^{\mathrm{T}}\hat{g}_{ref}-\mathbf{\epsilon}}{\hat{g}_{ref}^{\mathrm{T}}\hat{g}_{ref}}\hat{g}_{ref} $.

The optimization object of $\epsilon$-SOFT-GEM is:
\begin{equation}
\label{s_asoftgem_eq_1}
\begin{array}{ll}
\text{minimize}_{\widetilde{g}} & \quad \frac{1}{2}||\hat{g} - \widetilde{g}||_{2}^{2} \\
\text{subject to} & \quad \widetilde{g}^{\mathrm{T}}\hat{g}_{ref} \ge \mathbf{\epsilon}, \quad \epsilon \in [0, 1].
\end{array}
\end{equation}

Replacing $\widetilde{g}$ with $z$ and rewriting (\ref{s_asoftgem_eq_1}) yields:
\begin{equation}
\label{s_asoftgem_eq_2}
\begin{array}{ll}
\text{minimize}_{z} & \quad \frac{1}{2}z^{\mathrm{T}}z - \hat{g}^{\mathrm{T}}z  \\
\text{subject to} & \quad -z^{\mathrm{T}}\hat{g}_{ref} + \epsilon \le 0, \quad \epsilon \in [0, 1].
\end{array}
\end{equation}

Putting the cost function as well as the constraints in a single minimization problem, the dual optimization problem of (\ref{s_asoftgem_eq_2}) can be written as:
\begin{equation}
\label{s_asoftgem_eq_3}
\begin{array}{l}
\mathcal{L}(z, \alpha) = \frac{1}{2}z^{\mathrm{T}}z - \hat{g}^{\mathrm{T}}z - \alpha(z^{\mathrm{T}}\hat{g}_{ref} - \mathbf{\epsilon}),
\end{array}
\end{equation}
where $\alpha$ is a Lagrange multiplier.

The solution of (\ref{s_asoftgem_eq_3}) satisfies:
\begin{equation}
\label{s_asoftgem_eq_4_1}
\begin{array}{l}
\nabla_{z}\mathcal{L}(z, \alpha) = 0
\end{array}
\end{equation}
and
\begin{equation}
\label{s_asoftgem_eq_4_2}
\begin{array}{l}
\nabla_{\alpha}\mathcal{L}(z, \alpha) = 0.
\end{array}
\end{equation}

Consider (\ref{s_asoftgem_eq_3}) and (\ref{s_asoftgem_eq_4_1}),
\begin{equation}
\label{s_asoftgem_eq_4_5}
\begin{array}{l}
\nabla_{z}\mathcal{L}(z, \alpha) = z^{*} - \hat{g} - \alpha \hat{g}_{ref} = 0,
\end{array}
\end{equation}
the solution $z^{*}$ of (\ref{s_asoftgem_eq_4_5}) is:
\begin{equation}
\label{s_asoftgem_eq_4_5_2}
\begin{array}{l}
z^{*} = \hat{g} + \alpha \hat{g}_{ref}.
\end{array}
\end{equation}

Now, plug $z^{*}$ into equation (\ref{s_asoftgem_eq_3}) to obtain
\begin{equation}
\label{s_asoftgem_eq_6}
\begin{array}{ll}
\mathcal{L}(z^{*}, \alpha) &= \frac{1}{2}(\hat{g}^{\mathrm{T}} + \alpha \hat{g}_{ref}^{\mathrm{T}})(\hat{g} + \alpha \hat{g}_{ref})
- \hat{g}^{\mathrm{T}}(\hat{g} + \alpha \hat{g}_{ref}) - \alpha((\hat{g}^{\mathrm{T}} + \alpha \hat{g}_{ref}^{\mathrm{T}}) \hat{g}_{ref} - \mathbf{\epsilon}) \\
& = - \frac{1}{2}\hat{g}^{\mathrm{T}}\hat{g} - \alpha \hat{g}^{\mathrm{T}}\hat{g}_{ref} - \frac{1}{2}\alpha^{2}\hat{g}_{ref}^{\mathrm{T}}\hat{g}_{ref} + \alpha \mathbf{\epsilon}.
\end{array}
\end{equation}

Consider (\ref{s_asoftgem_eq_4_2}) and (\ref{s_asoftgem_eq_6}).
\begin{equation}
\label{s_asoftgem_eq_4_5_1}
\begin{array}{l}
\nabla_{\alpha}\mathcal{L}(z^{*}, \alpha) = - \hat{g}^{\mathrm{T}}\hat{g}_{ref} - \alpha \hat{g}_{ref}^{\mathrm{T}}\hat{g}_{ref} + \mathbf{\epsilon} = 0,
\end{array}
\end{equation}
the solution $\alpha^{*}$ of (\ref{s_asoftgem_eq_4_5_1}) is:
\begin{equation}
\label{s_asoftgem_eq_7}
\begin{array}{l}
\alpha^{*} = -\frac{\hat{g}^{\mathrm{T}}\hat{g}_{ref} - \mathbf{\epsilon}}{\hat{g}_{ref}^{\mathrm{T}}\hat{g}_{ref}}.
\end{array}
\end{equation}

Plugging $\alpha^{*}$ into equation (\ref{s_asoftgem_eq_4_5_2}), the SOFT-GEM update rule is obtained as:
\begin{equation}
\label{z*}
\begin{array}{l}
z^{*} = \hat{g} - \frac{\hat{g}^{\mathrm{T}}\hat{g}_{ref} - \mathbf{\epsilon}}{\hat{g}_{ref}^{\mathrm{T}}\hat{g}_{ref}}\hat{g}_{ref} = \widetilde{g}.
\end{array}
\end{equation}

\section{ALGORITHMS}
\label{appendix_soft_gem_alg}
The algorithm for $\epsilon$-SOFT-GEM and A-A-GEM is illustrated in \textbf{Algorithm} \ref{alg:soft_gem}, and the evaluation (EVAL) and update episodic memory (UPDATAEPSMEM) procedures are introduced from \cite{Chaudhry2018EfficientLL}.

\begin{algorithm}[H]
	\caption{Training and evaluation of $\epsilon$-SOFT-GEM and A-A-GEM on sequential data $\mathcal{D} = \{\mathcal{D}_{1}, ..., \mathcal{D}_{T}\} $}
	\label{alg:soft_gem}
	\begin{algorithmic}
		\Procedure {$\epsilon$-SOFT-GEM}{$\epsilon$, $f_{\theta}, \mathcal{D}^{train}, \mathcal{D}^{test}$}
		\State $\mathcal{M} \leftarrow \{\}$
		\State $A \leftarrow 0 \in \mathbb{R}^{\text{T} \times \text{T}}$
		\For {t = \{1, ..., T\}}
		\For {$(\mathbf{x}, y) \in \mathcal{D}_{t}^{train}$}
		\State $(\mathbf{x}_{ref}, y_{ref}) \sim \mathcal{M}$
		\State $g_{ref} \leftarrow \nabla_{\theta}l(f_{\theta}(\mathbf{x}_{ref}, t), y_{ref})$
		\State $g \leftarrow \nabla_{\theta}l(f_{\theta}(\mathbf{x}, t), y)$
		%\State $\hat{g} \leftarrow \frac{g}{|g|}$ , ${\hat{g}^{\text{T}}} \leftarrow \frac{g^{\text{T}}}{|g^{\text{T}}|}$,  $\hat{g}_{ref} \leftarrow \frac{g_{ref}}{|g_{ref}|}$ and ${\hat{g}_{ref}^{\text{T}}} \leftarrow \frac{g_{ref}^{\text{T}}}{|g_{ref}^{\text{T}}|}$ \Comment normalize $g$, $g_{ref}$, $g^{\text{T}}$ and ${g_{ref}^{\text{T}}}$
		\State $\hat{g} \leftarrow \frac{g}{|g|}$ ,$\hat{g}_{ref} \leftarrow \frac{g_{ref}}{|g_{ref}|}$ \Comment Normalize $g$ and $g_{ref}$ %  $g^{\text{T}}$ and ${g_{ref}^{\text{T}}}$
		\If {$\hat{g}^{T}\hat{g}_{ref} \ge 0$}
		\State $\widetilde{g} \leftarrow g$
		\Else
		% \State $\widetilde{g} \leftarrow g - \frac{{g^{'}}^{\text{T}}g_{ref}^{'} - \epsilon}{{g_{ref}^{'}}^{\text{T}}g_{ref}^{'}}g_{ref}$
		\If {$\epsilon$-SOFT-GEM}
		\State $\widetilde{g} \leftarrow \hat{g} - \frac{{{\hat{g}^{\text{T}}}}\hat{g}_{ref} - \epsilon}{{{\hat{g}_{ref}^{\text{T}}}}\hat{g}_{ref}}\hat{g}_{ref}$ \Comment $\epsilon$-SOFT-GEM update rule
		\ElsIf {A-A-GEM}
		\State $\widetilde{g} \leftarrow \frac{\hat{g} + \hat{g}_{ref}}{2}$ \Comment A-A-GEM update rule
		\EndIf
		\EndIf
		\State $\theta \leftarrow \theta - \alpha \widetilde{g}$
		\EndFor
		
		\State $M \leftarrow \mathrm{UPDATAEPSMEM}(\mathcal{M}, \mathcal{D}_{t}^{train}, T)$ \Comment Update the episodic memory
		\State $A_{t, :} \leftarrow \mathrm{EVAL}(f_{\theta}, \mathcal{D}^{test})$ \Comment Evaluate the model
		\EndFor \\
		\Return $f_{\theta}, A$
		\EndProcedure
	\end{algorithmic}
\end{algorithm}

\section{RESULTS}

\label{RESULTS}

\begin{table}[H]
	\scriptsize
	\begin{center}
		\caption{Models on Permuted MNIST}
		\label{tab:dataset_mnist_cifar_statistics_permuted_mnist}
		\begin{tabular}{|c|cccccc|}
			\hline
			\multirow{2}*{\textbf{Methods}} & \multicolumn{6}{c|}{\textbf{Permuted MNIST}}  \\
			\cline{2-7}
			& $A_{T}(\%)(\uparrow)$      &  $F_{T}(\downarrow)$ &  $a_{1}(\%)(\uparrow)$  & $a_{t}(\%)(\uparrow)$      &  $FWT(\%)(\uparrow)$ & $BWT(\%)(\uparrow)$        \\ \hline
			VAN             & $48.6(\pm 1.28)$ & $0.21(\pm 0.013)$ & $21.1(\pm 2.81)$ & $96.2(\pm 0.22)$ & $1.12(\pm 0.403)$ & $-50.68(\pm 1.342)$\\
			EWC             & $67.6(\pm 1.83)$ & $0.30(\pm 0.019)$ & $42.8(\pm 4.48)$ & $95.8(\pm 0.20)$ & $1.06(\pm 0.420)$ & $-29.84(\pm 1.906)$\\
			MAS             & $70.7(\pm 1.29)$ & $0.26(\pm 0.013)$ & $45.2(\pm 3.86)$ & $95.3(\pm 0.25)$ & $\bm{1.14(\pm 0.523)}$ & $-26.06(\pm 1.361)$\\
			RWALK           & $86.3(\pm 1.00)$ & $0.07(\pm 0.009)$ & $93.2(\pm 0.67)$ & $92.2(\pm 0.99)$ & $0.84(\pm 0.488)$ & $-6.94(\pm 0.935)$\\
			ER              & $90.6(\pm 0.07)$ & $0.04(\pm 0.001)$ & $91.6(\pm 0.45)$ & $94.2(\pm 0.15)$ & $0.94(\pm 0.159)$ & $-3.83(\pm 0.068)$\\
			A-GEM           & $89.8(\pm 0.32)$ & $0.07(\pm 0.003)$ & $84.6(\pm 0.42)$ & $\bm{96.0(\pm 0.28)}$ & $0.82(\pm 0.383)$ & $-6.60(\pm 0.338)$\\
			\textbf{A-A-GEM}      & $90.7(\pm 0.20)$ & $0.05(\pm 0.001)$ & $85.0(\pm 0.72)$ & $95.8(\pm 0.07)$ & $0.68(\pm 0.178)$ & $-5.09(\pm 0.190)$\\
			\textbf{$\epsilon$-SOFT-GEM}        & $91.3(\pm 0.11)$ & $0.05(\pm 0.001)$ & $87.2(\pm 0.32)$ & $95.1(\pm 0.28)$ & $0.81(\pm 0.414)$ & $-4.49(\pm 0.123)$\\
			PROGNN          & $\bm{93.5(\pm 0.10)}$ & $\bm{0(\pm 0)}$ & $\bm{96.2(\pm 0.27)}$ & $90.9(\pm 0.55)$ & $-0.07(\pm 0.817)$ & $\bm{0(\pm 0)}$\\
			\hline
		\end{tabular}
	\end{center}
\end{table}

\begin{table}[H]
	\scriptsize
	\begin{center}
		\caption{Models on Split CIFAR}
		\label{tab:dataset_mnist_cifar_statistics_cifar}
		\begin{tabular}{|c|cccccc|}
			\hline
			\multirow{2}*{\textbf{Methods}} & \multicolumn{6}{c|}{\textbf{Split CIFAR}}  \\
			\cline{2-7}
			& $A_{T}(\%)(\uparrow)$      &  $F_{T}(\downarrow)$ &  $a_{1}(\%)(\uparrow)$       & $a_{t}(\%)(\uparrow)$      &  $FWT(\%)(\uparrow)$ & $BWT(\%)(\uparrow)$        \\ \hline
			VAN             & $40.9(\pm 4.38)$ & $0.27(\pm 0.040)$ & $31.4(\pm 8.61)$ & $71.4(\pm 4.58)$ & $-0.06(\pm 1.570)$ & $-27.07(\pm 4.037)$\\
			EWC             & $41.8(\pm 4.02)$ & $0.28(\pm 0.041)$ & $26.6(\pm 6.94)$ & $75.8(\pm 4.19)$ & $-0.44(\pm 0.871)$ & $-27.5(\pm 4.179)$\\
			MAS             & $41.8(\pm 4.73)$ & $0.28(\pm 0.042)$ & $22.8(\pm 6.94)$ & $72.8(\pm 7.86)$ & $-0.17(\pm 0.611)$ & $-27.6(\pm 4.191)$\\
			RWALK           & $42.2(\pm 4.80)$ & $0.28(\pm 0.042)$ & $29.5(\pm 7.56)$ & $\bm{76.6(\pm 4.88)}$ & $-0.12(\pm 0.580)$ & $-27.7(\pm 4.307)$\\
			ER              & $45.0(\pm 2.27)$ & $0.11(\pm 0.009)$ & $47.4(\pm 8.94)$ & $54.8(\pm 9.04)$ & $-0.47(\pm 0.978)$ & $-10.6(\pm 0.821)$\\
			A-GEM           & $58.6(\pm 3.67)$ & $0.11(\pm 0.034)$ & $56.1(\pm 11.45)$ & $76.0(\pm 4.00)$ & $-0.16(\pm 1.118)$ & $-10.2(\pm 3.193)$\\
			\textbf{A-A-GEM}      & $62.1(\pm 1.97)$ & $0.06(\pm 0.010)$ & $61.1(\pm 8.67)$ & $69.8(\pm 6.30)$ & $-0.72(\pm 0.853)$ & $-3.02(\pm 1.295)$\\
			\textbf{$\epsilon$-SOFT-GEM}         & $\bm{63.9(\pm 1.53)}$ & $0.06(\pm 0.014)$ & $60.1(\pm 9.21)$ & $75.1(\pm 5.54)$ & $\bm{0.15(\pm 1.156)}$ & $-3.69(\pm 0.797)$\\
			PROGNN          & $59.6(\pm 0.92)$ & $\bm{0(\pm 0)}$ & $\bm{64.6(\pm 7.14)}$ & $65.8(\pm 5.45)$ & $0.01(\pm 0.525)$ & $\bm{0(\pm 0)}$\\
			\hline
		\end{tabular}
	\end{center}
\end{table}

\begin{table}[H]
	\scriptsize
	\begin{center}
		\caption{Models on Split CUB with one-hot. 'OoM' in the table means that the model fails to fit within the memory.}
		\label{tab:dataset_cub_statistics_tab_ohot}
		\begin{tabular}{|c|cccccc|}
			\hline
			\multirow{2}*{\textbf{Methods}} & \multicolumn{6}{c|}{\textbf{Split CUB}}  \\
			\cline{2-7}
			& $A_{T}(\%)(\uparrow)$      &  $F_{T}(\downarrow)$ &  $a_{1}(\%)(\uparrow)$      & $a_{t}(\%)(\uparrow)$      &  $FWT(\%)(\uparrow)$ & $BWT(\%)(\uparrow)$        \\ \hline
			VAN             & $53.4(\pm 0.76)$ & $0.12(\pm 0.030)$ & $34.5(\pm 7.66)$  & $64.9(\pm 7.00)$ & $\bm{0.42(\pm 0.93})$ & $-10.71(\pm 3.354)$\\
			EWC             & $54.3(\pm 1.73)$ & $0.11(\pm )0.015$ & $37.1(\pm 8.21)$  & $64.8(\pm 11.25)$ & $-0.09(\pm 1.04)$ & $-8.69(\pm 2.360)$\\
			PI              & $56.6(\pm 2.91)$ & $0.09(\pm )0.005$ & $40.2(\pm 7.35)$  & $69.3(\pm 10.36)$ & $-0.22(\pm 0.57)$ & $-6.41(\pm 1.125)$\\
			RWALK           & $54.5(\pm 1.50)$ & $0.11(\pm 0.016)$ & $34.8(\pm 8.35)$  & $70.0(\pm 7.58)$ & $-0.26(\pm 0.93)$ & $-9.26(\pm 2.434)$\\
			% ER              & $(\pm )$ & $(\pm )$ & $(\pm )$ \\
			A-GEM           & $61.1(\pm 3.12)$ & $0.08(\pm 0.012)$ & $56.6(\pm 5.81)$ & $69.6(\pm 6.83)$ & $0.09(\pm 0.33)$ & $-4.19(\pm 2.384)$\\
			\textbf{A-A-GEM}      & $66.7(\pm 2.40)$ & $\bm{0.02(\pm 0.015)}$ & $\bm{73.9(\pm 5.50)}$ & $54.7(\pm 10.80)$ & $-0.25(\pm 0.34)$ & $\bm{21.70(\pm 2.083)}$\\
			\textbf{$\epsilon$-SOFT-GEM}         & $\bm{75.6(\pm 2.00)}$ & $0.03(\pm 0.009)$ & $73.3(\pm 4.95)$ & $\bm{70.7(\pm 5.52)}$ & $0.28(\pm 0.98)$ & $13.79(\pm 2.979)$\\
			PROGNN          & OoM & OoM & OoM  & OoM & OoM & OoM \\
			\hline
		\end{tabular}
	\end{center}
\end{table}

\begin{table}[H]
	\scriptsize
	\begin{center}
		\caption{Models on Split CUB with joint embedding. 'OoM' in the table means that the model fails to fit within the memory.}
		\label{tab:dataset_cub_statistics_tab_je}
		\begin{tabular}{|c|cccccc|}
			\hline
			\multirow{2}*{\textbf{Methods}} & \multicolumn{6}{c|}{\textbf{Split CUB(-JE)}}  \\
			\cline{2-7}
			& $A_{T}(\%)(\uparrow)$      &  $F_{T}(\downarrow)$ &  $a_{1}(\%)(\uparrow)$      & $a_{t}(\%)(\uparrow)$      &  $FWT(\%)(\uparrow)$ & $BWT(\%)(\uparrow)$        \\ \hline
			VAN             & $67.7(\pm 5.40)$ & $0.10(\pm 0.057)$ & $58.1(\pm 10.39)$ & $79.3(\pm 5.34)$ & $32.47(\pm 2.867)$ & $-5.75(\pm 5.758)$\\
			EWC             & $67.5(\pm 4.61)$ & $0.10(\pm 0.043)$ & $57.6(\pm 9.41)$ & $79.6(\pm 4.99)$ & $31.6(\pm 2.530)$ & $-5.41(\pm 4.814)$\\
			PI              & $65.8(\pm 5.71)$ & $0.11(\pm 0.068)$ & $54.6(\pm 8.80)$ & $78.1(\pm 4.43)$ & $32.19(\pm 2.616)$ & $-7.24(\pm 6.758)$\\
			RWALK           & $67.1(\pm 4.20)$ & $0.10(\pm 0.041)$ & $58.1(\pm 9.10)$ & $79.5(\pm 5.60)$ & $32.04(\pm 1.976)$ & $-5.77(\pm 3.464)$\\
			% ER              & $(\pm )$ & $(\pm )$ & $(\pm )$ \\
			A-GEM           & $71.1(\pm 3.59)$ & $0.08(\pm 0.020)$ & $64.3(\pm 10.27)$  & $81.2(\pm 6.99)$ & $34.92(\pm 2.585)$ & $-4.23(\pm 1.605)$\\
			\textbf{A-A-GEM}      & $70.1(\pm 2.40)$ & $0.08(\pm 0.028)$ & $68.2(\pm 5.73)$ & $77.0(\pm 4.58)$ & $35.61(\pm 2.697)$ & $-2.80(\pm 2.954)$\\
			\textbf{$\epsilon$-SOFT-GEM}         & $\bm{79.8(\pm 2.73)}$ & $\bm{0.03(\pm 0.021)}$ & $\bm{80.2(\pm 6.25)}$ & $\bm{82.7(\pm 4.55)}$ & $\bm{36.71(\pm 2.432)}$ & $\bm{6.54(\pm 2.639)}$\\
			PROGNN          & OoM & OoM & OoM  & OoM & OoM & OoM \\
			\hline
		\end{tabular}
	\end{center}
\end{table}

\begin{table}[H]
	\scriptsize
	\begin{center}
		\caption{Models on Split AWA with one-hot. 'OoM' in the table means that the model fails to fit within the memory.}
		\label{tab:dataset_awa_statistics_tab_ohot}
		\begin{tabular}{|c|cccccc|}
			\hline
			\multirow{2}*{\textbf{Methods}} & \multicolumn{6}{c|}{\textbf{Split AWA}}  \\
			\cline{2-7}
			& $A_{T}(\%)(\uparrow)$      &  $F_{T}(\downarrow)$ &  $a_{1}(\%)(\uparrow)$   & $a_{t}(\%)(\uparrow)$      &  $FWT(\%)(\uparrow)$ & $BWT(\%)(\uparrow)$        \\ \hline
			VAN             & $29.4(\pm 2.65)$ & $0.03(\pm 0.008)$ & $26.8(\pm 5.01)$ & $34.2(\pm 8.60)$ & $-0.56(\pm 1.218)$ & $-2.09(\pm 0.781)$\\
			EWC             & $33.7(\pm 3.52)$ & $0.07(\pm 0.017)$ & $27.9(\pm 4.16)$ & $43.3(\pm 7.64)$ & $-0.68(\pm 0.630)$ & $-5.83(\pm 1.796)$\\
			PI              & $33.5(\pm 3.65)$ & $0.07(\pm 0.017)$ & $28.4(\pm 4.39)$ & $43.6(\pm 7.64)$ & $-0.52(\pm 0.641)$ & $-6.33(\pm 1.575)$\\
			RWALK           & $33.7(\pm 3.17)$ & $0.07(\pm 0.020)$ & $29.1(\pm 5.07)$  & $43.5(\pm 6.81)$ & $-0.78(\pm 0.459)$ & $-6.05(\pm 2.049)$\\
			A-GEM           & $43.0(\pm 3.24)$ & $0.05(\pm 0.015)$ & $42.8(\pm 7.53)$ & $45.0(\pm 5.58)$ & $0.33(\pm 1.088)$ & $-0.97(\pm 0.856)$\\
			\textbf{A-A-GEM}      & $46.3(\pm 3.64)$ & $0.02(\pm 0.027)$ & $47.3(\pm 9.36)$ & $39.2(\pm 7.58)$ & $0.91(\pm 1.443)$ & $5.19(\pm 2.542)$\\
			\textbf{$\epsilon$-SOFT-GEM}         & $\bm{55.3(\pm 3.57)}$ & $\bm{0.01(\pm 0.028)}$ & $ \bm{53.4(\pm 8.88)}$  & $\bm{45.3(\pm 7.99)}$ & $\bm{1.38(\pm 1.453)}$ & $\bm{10.65(\pm 2.051)}$\\
			PROGNN          & OoM & OoM & OoM & OoM & OoM & OoM \\
			\hline
		\end{tabular}
	\end{center}
\end{table}

\begin{table}[H]
	\scriptsize
	\begin{center}
		\caption{Models on Split AWA with joint embedding. 'OoM' in the table means that the model fails to fit within the memory.}
		\label{tab:dataset_awa_statistics_tab_je}
		\begin{tabular}{|c|cccccc|}
			\hline
			\multirow{2}*{\textbf{Methods}} & \multicolumn{6}{c|}{\textbf{Split AWA (-JE)}}  \\
			\cline{2-7}
			& $A_{T}(\%)(\uparrow)$      &  $F_{T}(\downarrow)$ &  $a_{1}(\%)(\uparrow)$ & $a_{t}(\%)(\uparrow)$      &  $FWT(\%)(\uparrow)$ & $BWT(\%)(\uparrow)$        \\ \hline
			VAN             & $44.3(\pm 3.05)$ & $0.05(\pm 0.015)$ & $39.9(\pm 2.30)$ & $51.7(\pm 3.83)$ & $11.65(\pm 1.677)$ & $-0.93(\pm 1.575)$\\
			EWC             & $45.1(\pm 2.61)$ & $0.04(\pm 0.011)$ & $40.5(\pm 3.45)$ & $50.5(\pm 5.07)$ & $11.81(\pm 1.930)$ & $0.19(\pm 0.992)$\\
			PI              & $43.7(\pm 3.53)$ & $0.05(\pm 0.015)$ & $40.7(\pm 3.98)$ & $48.6(\pm 3.78)$ & $12.26(\pm 1.892)$ & $0.03(\pm 1.024)$\\
			RWALK           & $44.0(\pm 2.75)$ & $0.05(\pm 0.012)$ & $39.9(\pm 2.68)$  & $44.0(\pm 3.53)$ & $12.00(\pm 1.938)$ & $-1.42(\pm 1.537)$\\
			% ER              & $(\pm )$ & $(\pm )$ & $(\pm )$ \\
			A-GEM           & $48.4(\pm 2.58)$ & $0.03(\pm 0.007)$ & $49.6(\pm 7.65)$ & $\bm{49.7(\pm 4.54)}$ & $15.19(\pm 1.591)$ & $2.60(\pm 1.191)$\\
			\textbf{A-A-GEM}      & $47.2(\pm 2.20)$ & $\bm{0.02(\pm 0.008)}$ & $48.0(\pm 5.60)$ & $48.8(\pm 3.16)$ & $14.68(\pm 1.035)$ & $4.53(\pm 1.217)$\\
			\textbf{$\epsilon$-SOFT-GEM}         & $\bm{53.5(\pm 5.07)}$ & $0.04(\pm 0.038)$ & $\bm{54.6(\pm 7.73)}$  & $48.7(\pm 6.81)$ & $\bm{20.22(\pm 1.698)}$ & $\bm{6.30(\pm 3.682)}$\\
			PROGNN          & OoM & OoM & OoM & OoM & OoM & OoM \\
			\hline
		\end{tabular}
	\end{center}
\end{table}

\begin{table}[H]
	\footnotesize
	\begin{center}
		\caption{Permuted MNIST  with varying episodic memory size}
		\label{tab:dataset_permuted_mnist_episodic_memory_statistics_mnist}
		\begin{tabular}{|c|c|c|c|c|}
			\hline
			\multicolumn{2}{|c|}{\textbf{Methods}}  & A-GEM             & \textbf{A-A-GEM}  & \textbf{$\epsilon$-SOFT-GEM}\\\hline
			
			\multirow{3}*{$A_{T}(\%)(\uparrow)$}    & 25      & $89.6(\pm 0.32)$  & $90.7(\pm 0.20)$       & $\bm{91.3(\pm 0.11)}$ \\
			
			& 50      & $89.5(\pm 0.04)$  & $92.0(\pm 0.09)$       & $\bm{92.3(\pm 0.15)}$  \\
			
			& 100     & $89.8(\pm 0.45)$  & $92.7(\pm 0.11)$       & $\bm{92.9(\pm 0.08)}$ \\  \hline
			\multirow{3}*{$a_{1}(\%)(\uparrow)$}    & 25      & $84.6(\pm 0.43)$  & $85.0(\pm 0.72)$       & $\bm{87.2(\pm 0.32)}$ \\
			
			& 50      & $85.7(\pm 1.05)$  & $87.5(\pm 0.33)$       & $\bm{89.3(\pm 0.60)}$ \\
			
			& 100      & $87.4(\pm 1.17)$  & $89.6(\pm 0.32)$       & $\bm{90.7(\pm 0.34)}$ \\ \hline
			\multirow{3}*{$a_{t}(\%)(\uparrow)$}    & 25      & $\bm{96.0(\pm 0.25)}$  & $95.8(\pm 0.07)$       & $95.1(\pm 0.28)$ \\
			
			& 50      & $95.6(\pm 0.19)$  & $\bm{95.8(\pm 0.17)}$       & $95.3(\pm 0.18)$ \\
			
			& 100     & $94.2(\pm 0.47)$  & $\bm{95.3(\pm 0.18)}$       & $94.4(\pm 0.50)$ \\ \hline
			
			\multirow{3}*{$F_{T}(\downarrow)$}        & 25      & $0.07(\pm 0.003)$ & $0.05(\pm 0.001)$      & $\bm{0.05(\pm 0.001)}$ \\
			
			& 50      & $0.07(\pm 0.001)$ & $0.04(\pm 0.001)$      & $\bm{0.04(\pm 0.001)}$ \\
			
			& 100     & $0.06(\pm 0.004)$ & $0.03(\pm 0.001)$      & $\bm{0.03(\pm 0.001)}$ \\
			
			\hline
			
		\end{tabular}
	\end{center}
\end{table}

\begin{table}[H]
	\footnotesize
	\begin{center}
		\caption{Split CIFAR with varying episodic memory size}
		\label{tab:dataset_permuted_mnist_episodic_memory_statistics_cifar}
		\begin{tabular}{|c|c|c|c|c|}
			
			\hline
			\multicolumn{2}{|c|}{\textbf{Methods}}  & A-GEM             & \textbf{A-A-GEM}  & \textbf{$\epsilon$-SOFT-GEM} \\\hline
			
			\multirow{3}*{$A_{T}(\%)(\uparrow)$}    & 13      & $58.6(\pm 3.67)$  & $62.1(\pm 1.97)$       & $\bm{63.9(\pm 1.53)}$ \\
			
			& 25      & $60.8(\pm 1.47)$  & $65.0(\pm 1.09)$       & $\bm{66.4(\pm 2.07)}$ \\
			
			& 50      & $62.0(\pm 2.00)$  & $67.2(\pm 1.95)$       & $\bm{69.3(\pm 1.76)}$ \\  \hline
			
			\multirow{3}*{$a_{1}(\%)(\uparrow)$}    & 13      & $56.1(\pm 11.45)$  & $\bm{61.1(\pm 8.67)}$       & $60.1(\pm 9.21)$ \\
			
			& 25      & $61.3(\pm 7.95)$  & $\bm{66.2(\pm 6.47)}$       & $64.0(\pm 7.03)$ \\
			
			& 50      & $60.4(\pm 8.86)$  & $67.0(\pm 8.22)$       & $\bm{68.5(\pm 5.05)}$ \\ \hline
			\multirow{3}*{$a_{t}(\%)(\uparrow)$}    & 13      & $\bm{76.0(\pm 4.00)}$  & $69.8(\pm 6.30)$       & $75.1(\pm 6.54)$ \\
			
			& 25      & $74.6(\pm 7.94)$  & $74.2(\pm 4.17)$       & $\bm{74.9(\pm 8.58)}$ \\
			
			& 50      & $73.4(\pm 4.66)$  & $71.7(\pm 6.76)$       & $\bm{73.4(\pm 3.97)}$ \\ \hline
			
			\multirow{3}*{$F_{T}(\downarrow)$}        & 13      & $0.11(\pm 0.034)$ & $0.06(\pm 0.010)$      & $\bm{0.06(\pm 0.014)}$ \\
			
			& 25      & $0.10(\pm 0.013)$ & $\bm{0.04(\pm 0.008)}$ & $0.05(\pm 0.020)$ \\
			
			& 50      & $0.09(\pm 0.023)$ & $\bm{0.03(\pm 0.010)}$ & $0.03(\pm 0.019)$ \\
			
			\hline
		\end{tabular}
	\end{center}
\end{table}

\end{document}